\documentclass[letterpaper, 10 pt, conference]{ieeeconf}
\usepackage{commands}

\IEEEoverridecommandlockouts                     
\usepackage{graphicx}
\usepackage{amsmath}
\usepackage{amssymb}
\usepackage{booktabs}
\usepackage{transparent}
\usepackage{nicematrix}
\usepackage{pifont}
\usepackage{stfloats}
\usepackage{wrapfig}
\usepackage{ifthen}
\usepackage{graphicx}
\usepackage{tikz}
\usetikzlibrary{decorations.pathreplacing, calligraphy}
\usetikzlibrary{arrows}

\newcommand\num{\color{red}XXX\xspace}
\newcommand{\cmark}{\ding{51}}%
\newcommand{\xmark}{\ding{55}}%
\newcommand\header[1]{\textbf{#1}}
\newcommand{\epsbf}{\boldsymbol{\epsilon}}
\newcommand{\M}{\boldsymbol{M}}
\newcommand\Lcal{\mathcal{L}}
\newcommand\trilanes{{RaceLane}\xspace}

\usepackage[capitalize]{cleveref}
\crefname{section}{Sec.}{Secs.}
\Crefname{section}{Section}{Sections}
\Crefname{table}{Table}{Tables}
\crefname{table}{Tab.}{Tabs.}
\Crefname{figure}{Figure}{Figures}
\crefname{figure}{Fig.}{Figs.}
\Crefname{appendix}{Appendix}{Appendices}
\crefname{appendix}{App.}{Apps.}
\Crefname{algorithm}{Algorithm}{Algorithms}
\crefname{algorithm}{Alg.}{Algs.}

\newcommand\bestcolor{gray!40}
\newcommand\secondbestcolor{gray!15}

\newcommand\ourmethod{\textsc{EnsembleLanes}\xspace}

\title{\LARGE \bf
3D Lane Detection with Odometry for High-Speed Vehicle Racing
}

\author{
Omoruyi Atheka$^{1}$, John Subosits$^2$, and Marcus Greiff$^2$
\thanks{$^*$The work of O. Atheka was done while at TRI.}
\thanks{$^1$Massachusetts Institute of Technology (MIT),\hspace{-1pt}  Cambridge,\hspace{-1pt} MA,\hspace{-1pt} USA.}
\thanks{$^2$Toyota Research Institute (TRI), 94022 Los Altos, CA, USA. Email of corresponding author: \texttt{marcus.greiff@tri.global}.}
}

\begin{document}

\maketitle
\thispagestyle{empty}
\pagestyle{empty}

\begin{abstract}
Lane boundary detection is a critical component in autonomous driving systems and has been rigorously studied in regular driving scenarios. However, it is less explored in vehicle racing, where the car moves at higher speeds across more extreme road geometries. To study this problem, we introduce a new dataset for 3D lane detection in racing, featuring >$250$k images from multiple camera feeds and inertial measurements taken with a Lexus LC 500 driving on a closed circuit. With this dataset, we compare various approaches to 3D lane detection and propose modifications that permit frames to be processed at rates of almost 300Hz while retaining high predictive performance in the racing application. This facilitates a multi-camera ensemble approach that is validated on hardware. We show that sensing modalities such as inertial measurements can be leveraged for pre-integration to regress road geometries over both cameras and time, yielding improvements in key metrics. Compared to methods such as BevLaneDet, adding odometry and ensemble predictions improves the F1 score by 3 points and reduces near-vehicle mean absolute errors (MAEs) by $>30 \%$. We show F1 scores $>$0.9 and lateral MAEs of $<$0.18m in vehicle deployments.
\end{abstract}    
\section{Introduction}\label{sec:intro}

Accurate and fast detection of lane boundaries is essential for autonomous vehicles (AVs) to ensure safe and robust navigation. Such methods provide AVs with useful information for localization~\cite{du2016vision,berntorp2022bayesian}, motion planning~\cite{dallas2023hierarchical,lew2024risk}, and road mapping~\cite{berntorp2024framework}. Due to its great practical utility, the problem has been rigorously studied for autonomous driving (AD) applications. However, it is largely unexplored for vehicle racing~\cite{ma2024monocular}, where the car drives highly dynamically over extreme road geometries. 
Road courses used for racing share characteristics such as sharp curves, hills, and frequent absence of road shoulders with rural roads, where the majority of traffic fatalities occur in the United States~\cite{safety2004federal}, suggesting that common approaches may perform well in both domains.

\begin{figure}[t!]
    \centering
    %\hspace*{-2.2cm}\resizebox{10cm}{!}{\input{tikz/flagship}
    \includegraphics[width=0.95\columnwidth]{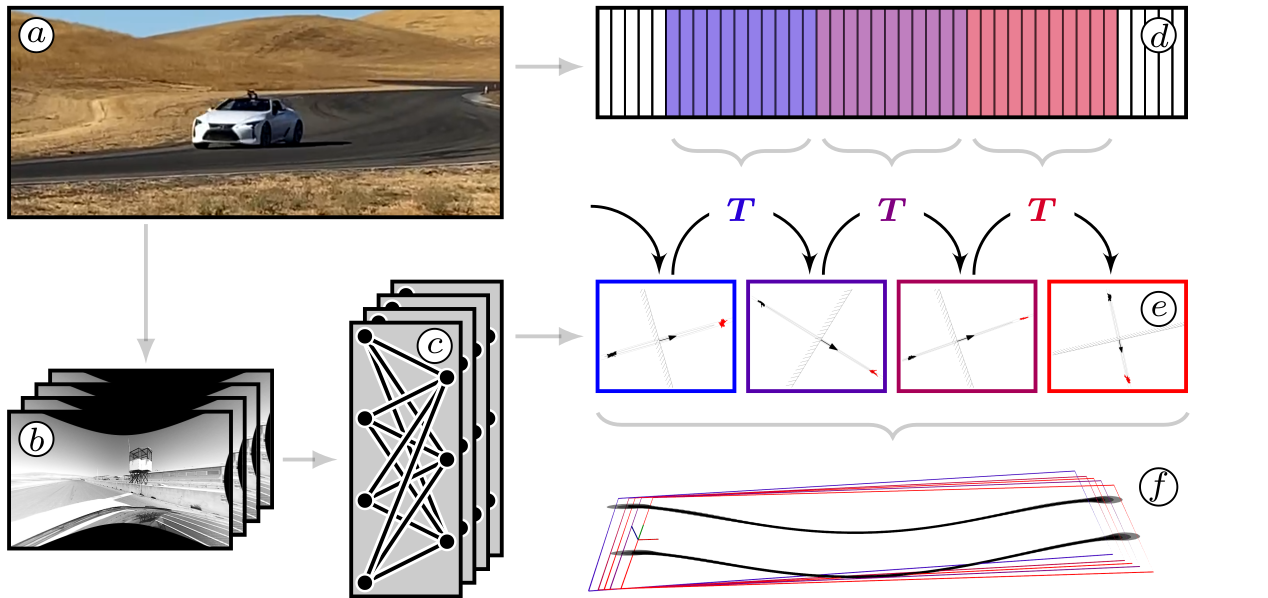}
    \vspace*{-12pt}
    
    \caption{Proposed method for 3D lane detection. (a) Distorted images are sampled in the car and (b) unwarped before (c) being passed into an ensemble of monocular 3D lane detection networks. Additionally, (d) IMU and CAN bus measurements are used to compute a buffer of transforms, which are (e) combined into a chain of transforms for temporal linking of clusters. This is used to (f) regress a parametric representation of the road.}
    \label{fig:flagship}
\end{figure}

Classical lane detection methods segment the lane boundaries in the camera's perspective utilizing an Inverse Perspective Mapping (IPM) under a flat ground assumption ~\cite{aly2008real,borkar2011novel,deusch2012random,hur2013multi,neven2018towards, huang2023anchor3dlane}, which is less useful for control and planning purposes if the road elevation is unknown or uncertain~\cite{wang2023bev}. 3D lane detection techniques have been proposed to infer the 3D geometry of lanes directly~\cite{guo2020gen,luo2023latr,pittner2024lanecpp}. Recent methods using Bird's Eye View (BEV) representations combined with learned 3D features have shown promising results. However, such methods are typically trained on monocular image data from urban environments~\cite{huang2020apollo,yan2022once}, which are not well suited for feature-sparse conditions and extreme road geometries encountered in the racing application (see~\cref{fig:flagship}.a). For dynamic driving on racetracks, the 3D lane detectors must be able to predict and potentially cluster 3D lane predictions at high rates. This is a significant constraint, as methods are based on deep learning techniques~\cite{luo2023latr,chen2022persformer,pittner2024lanecpp} in contrast to the (often simpler and faster) 2D detectors~\cite{aly2008real,borkar2011novel,deusch2012random,hur2013multi}. Finally, the cited works predict the lanes in each input image independently~\cite{guo2020gen,luo2023latr,chen2022persformer,pittner2024lanecpp,yan2022once,huang2020apollo}. Recent 3D lane detectors learn the geometric relationship between consecutively sampled images~\cite{bai2024curveformer,huang2023anchor3dlane}, but we hypothesize that performance may be improved further with signals that are ubiquitous in modern vehicles, such as speedometers or inertial measurements. 

To address 1) the lack of research on dynamic driving, 2) the inference time of existing 3D lane detectors, and 3) the under-utilization of odometry information,  we propose \ourmethod: a lane detection methodology developed for vehicle racing leveraging ensemble predictions (see Fig.~\ref{fig:flagship}). The method is agnostic to specific 3D lane detectors and can be implemented with~\cite{wang2023bev,luo2023latr,chen2022persformer,pittner2024lanecpp}, operates on synchronous camera feeds, and improves performance by fusing multiple independent predictions across time and across cameras using signals such as IMU and wheel odometry. %We optimize existing monocular detectors for racing. For the best-performing method, we show that the proposed ensemble method is tractable for real-time use and further improves key metrics. 
We also introduce \trilanes: the first benchmark 3D lane detection dataset for vehicle racing, comprising 250K images from the Thunderhill Raceway in California, annotated with lane boundaries, transforms, and inertial information.

Our main contributions are:
    
\begin{itemize}
    \item A dataset for supervised training of 3D lane detectors in autonomous racing, containing transforms and labeled data from a car driving at speed on a race track.

    \item A method of fusing information from asynchronous camera feeds using odometry, resulting in improved 3D accuracy and classification scores in 3D lane detection.
\end{itemize}

We emphasize that the optimizations of the various detector methods listed in Sec.~\ref{sec:adaptationsforracing} are important when using existing driving detectors in the racing application, where predictions are typically needed at higher rates than in regular driving. Unlike \ourmethod, these optimizations are model-specific and are therefore a lesser contribution of the paper.
\section{Related Work}\label{sec:review}

In racing, assuming that the road is planar and using an IPM may result in poor performance at longer distances~\cite{chen2023efficient}, motivating the use of 3D lane detectors. The pioneering GenLaneNet~\cite{garnett20193d}, later extended in~\cite{efrat20203d}, leverages a Spatial Transform Network (STN)~\cite{jaderberg2015spatial} to map features into a BEV perspective, proposing an end-to-end 3D lane detector. This spurred related work in~\cite{guo2020gen,chen2022persformer,li2022reconstruct,luo2023latr,wang2023bev,efrat20203d,pittner2024lanecpp,bai2023curveformer,bai2024curveformer,zheng2024pvalane,huang2023anchor3dlane}. A subset of these methods use an IPM internally~\cite{garnett20193d, guo2020gen, reiher2020sim2real,efrat20203d}. Others map 2D features to the BEV space using Multi-layer Perceptrons (MLPs)~\cite{pan2020cross, wang2023bev}, leverage depth predictions~\cite{pittner2024lanecpp}, use depth estimates~\cite{yan2022once}, or utilize transformer architectures~\cite{chen2022persformer,li2022bevformer,li2022reconstruct,luo2023latr}. Among these, PersFormer~\cite{chen2022persformer} has emerged as a popular anchor-based method, improved upon in LATR~\cite{luo2023latr} by removing surrogate representations, thereby achieving better performance and representing the state-of-the-art in transformer-based methods. There are analogs of the parametric methods~\cite{tabelini2021polylanenet} in the transformer-based CurveFormer~\cite{bai2023curveformer,bai2024curveformer} and Anchor3DLane~\cite{huang2023anchor3dlane}, which directly predict polynomial coefficients in 3D. Other approaches, such as SALAD~\cite{yan2022once}, combine 2D lane segmentation with monocular depth estimation~\cite{godard2019digging} to address the flat ground assumptions of the IPM. However, this is inaccurate at long distances~\cite{ma2024monocular} and adds computational complexity. As inference speed is key in vehicle racing, the MLP-based Bev-LaneDet~\cite{wang2023bev} is a fast and accurate alternative. It is not anchor-based, which may be advantageous when considering more extreme racetrack geometries. Recent methods, such as~\cite{pittner2024lanecpp,pittner2025sparselanestp}, showed marginal improvements over~\cite{wang2023bev} by using geometric priors. These results were not reproducible, and Bev-LaneDet remains competitive for regular driving.

To improve performance, some methods, such as PETRv2~\cite{liu2023petrv2}, combine vision with other sensing modalities~\cite{caltagirone2019lidar,bai2018deep,zhang2021channel,bai2022transfusion,zhao2024advancements3dlanedetection}. As depth information is limited in the monocular image data, LiDAR point clouds are commonly used for this purpose~\cite{caltagirone2019lidar,zhao2024advancements3dlanedetection,li2024sparsefusion}. This can be done using a depth estimate from the LiDAR to inform the IPM of a 2D lane detector~\cite{bai2018deep} (analogous to SALAD~\cite{yan2022once}); by including cross-model knowledge transfer through LiDAR~\cite{zhao2024lanecmkt}; or by directly convolving the LiDAR point clouds with image features~\cite{zhang2021channel,liu2023petrv2}. However, ubiquitous sensing modalities such as cheap IMU sensors may improve detections at a comparatively low cost.

Inertial information is relatively unexplored for 3D lane detection, which is typically treated as a static problem~\cite{yan2022once,wang2023bev,chen2022persformer}. Recently, extensions of CurveFormer~\cite{bai2023curveformer} in~\cite{bai2024curveformer} and methods such as Anchor3DLane-T~\cite{huang2023anchor3dlane} have broken from this tradition and shown encouraging performance when learning transforms relating features in consecutive frames (without IMU data). This works when driving at constant speeds, but may require stronger priors in the racing application. Odometry from the vehicle and inertial measurements have not yet been considered in state-of-the-art 3D lane predictors~\cite{ma2024monocular}. Despite not appearing in existing benchmarks and datasets, such measurements are ubiquitous in practice, and we hypothesize that leveraging ideas analogous to on-manifold pre-integration~\cite{forster2015imu,forster2016manifold} developed for visual-inertial SLAM~\cite{campos2021orb} can improve 3D lane detection in racing, and perhaps in driving more generally.
\section{Datasets}\label{sec:data}

3D lane detection datasets typically include RGB images with front forward-facing cameras and associated labels expressed as 3D points in a vehicle-fixed frame~\cite{huang2020apollo,xu2020curvelane,chen2022persformer,yan2022once}. In derivative works, 3D lane detection is most often treated as a static problem -- predicting the road geometry from single monocular images~\cite{wang2023bev,pittner2024lanecpp}. Consequently, relevant benchmarks do not contain ubiquitous vehicle data such as IMU measurements or wheel speeds. There are driving datasets with IMU measurements, such as KITTI~\cite{geiger2013vision} and nuScenes~\cite{caesar2020nuscenes}, but these do not come with 3D lane labels.

To study 3D lane detection in racing, we provide the \trilanes, comprising four monochromatic 600x480 camera feeds at 20Hz, IMU data at 500Hz, and wheel odometry data at 62.5Hz. It comprises recordings of real data from the Thunderhill Raceway in California, and includes 38 miles of driving with over 250K annotated images in different lighting conditions and seasons. The data is partitioned into an 80/20 split of 13 ``runs'' (laps), as {train} and {validation} splits. As we only have access to one racetrack, we include six runs collected at a different testing occasion as a {hold-out} split.

\begin{figure}[t!]
\centering
\includegraphics[width=0.95\columnwidth]{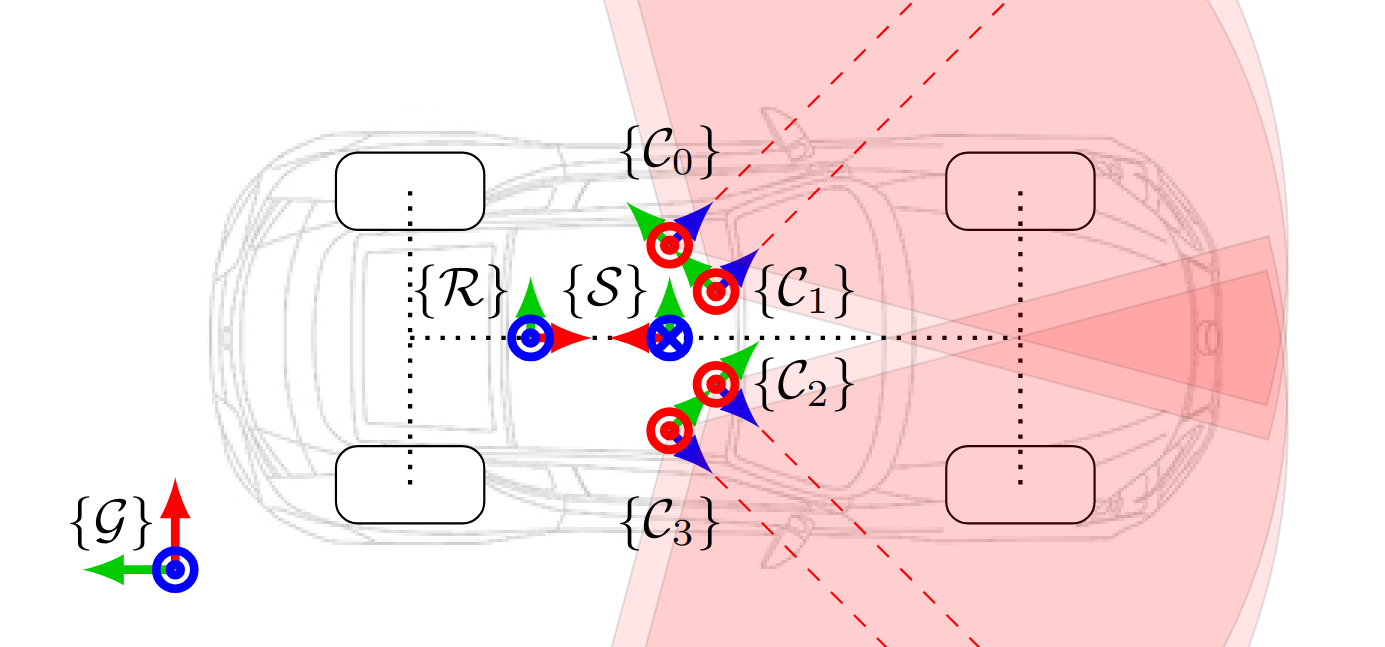}
\vspace{-10pt}

\caption{Top view of car with the coordinate frames used in \trilanes. All frames are vehicle-fixed, except for the global frame $\{\Gcal\}$. The $x|y|z$ directions are shown in the colors $r|g|b$. Camera FOVs are shown in red.}
\label{fig:cargeometry}
\end{figure}

 The sensors are mounted on top of the vehicle (see~\cref{fig:cargeometry}) and configured in various coordinate frames. Here, $\{\Gcal\}$ is a global ENU frame, $\{\mathcal{R}\}$ is a road-frame (origin on the road's surface), $\{\Scal\}$ is a sensor frame in which the IMU is mounted, and $\{\Ccal_i\}$ are camera frames. All frames are vehicle-fixed and differ by static transforms except for the global frame $\{\Gcal\}$. To generate labels, locally visible 3D lane labels by projecting sparse point clouds of the lane boundaries to image space using vehicle poses computed using an OxTS inertial navigation system~\cite{oxts2024}. Additional details on the dataset creation, formatting, and location are
 % provided here~\cite{atekha2026extended}.
 in the appendix.

\newcommand\bev{\textsc{Bev-LaneDet}\xspace}

\begin{figure*}[t!]
    \centering
    \hspace*{-20pt}%\resizebox{\textwidth}{!}{\input{tikz/method_second}}
    
\includegraphics[width=\textwidth]{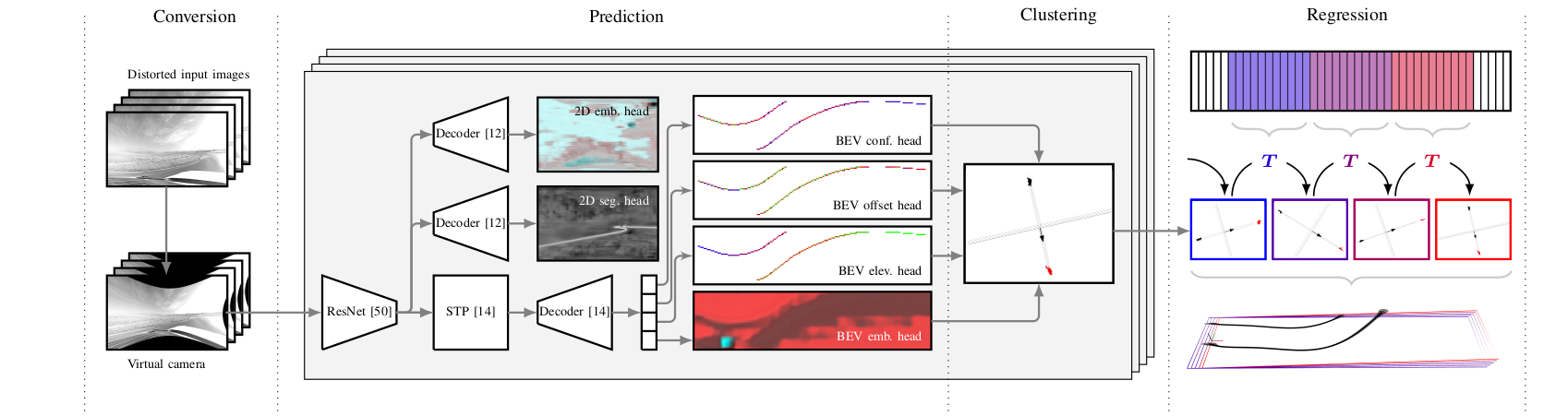}

    \vspace{-10pt}
    \caption{Illustration of \ourmethod with \bev for the monocular prediction in the ensemble, using odometry to regress the lanes over cameras and time (blue to red). Each gray box is a model with a clustering algorithm, and the ensemble of models is used to generate a regressed output.}
    \label{fig:networkarcitechture}
\end{figure*}
\section{Methodology}\label{sec:method}

To infer the lane boundaries in the racing context using \trilanes, we propose the \ourmethod method, which uses inertial odometry and is illustrated in~\cref{fig:networkarcitechture}. It consists of a neural network (NN) 3D lane predictor, a clustering algorithm, and a regressor. For the NN predictor, we primarily focus on the \bev architecture in~\cite{wang2023bev}, which in turn builds on a larger body of work in ~\cite{he2016deep,neven2018towards,garnett20193d,guo2020gen}. While not limited to this predictor, we introduce it in~\cref{sec:method:3dlane} to facilitate discussions on adaptations for racing (see~\cref{sec:method:adaptations}), defining new input/output spaces, discussing clustering methods, and other optimizations. The resulting models are used to explore regression of parametric lane boundaries to support multiple asynchronous camera feeds and in late fusion with IMU and velocity data (see~\cref{sec:method:regression}).

\begin{figure}[t!]
    \centering

    \tikzset{
        bracestyle/.style = {decorate,
        decoration={calligraphic brace, amplitude=4pt,
        raise=1pt, mirror},% for mirroring of brace
        very thick,
        pen colour=black}
    }
    \begin{tikzpicture}
        \node[inner sep=0pt] (image) at (0,0)
    {\includegraphics[width=.9\columnwidth]{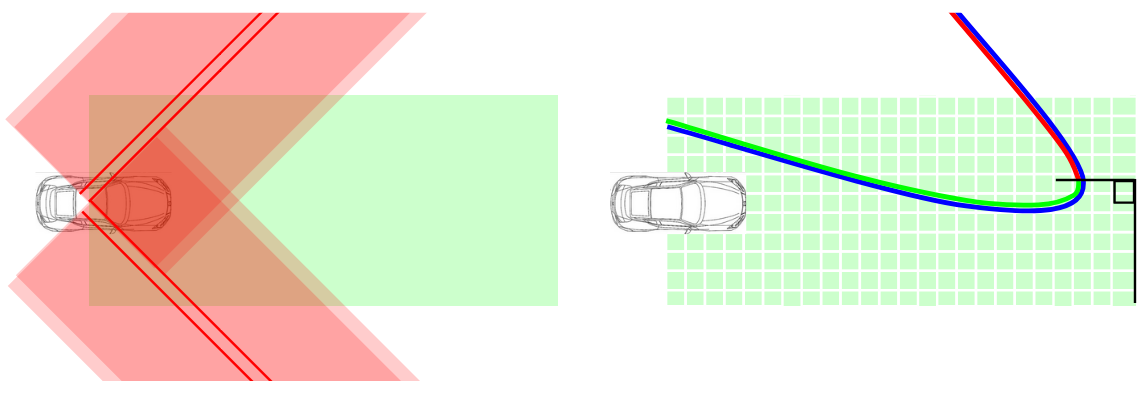}};
    \draw[bracestyle] (0.6,-0.71) -- node[below=2mm] {\scalebox{.7}{$M_{\mathrm{out}}$}} (3.66,-0.71);
    \draw[bracestyle] (3.68,-0.7) -- node[right=1mm] {\scalebox{.7}{$N_{\mathrm{out}}$}} (3.68,0.65);

    \node[draw=black] (cell) at (1.2,1.05) {\scalebox{.7}{BEV cell}};
    \draw[] (cell.south) -- (1.68, 0.62);
    \end{tikzpicture}
    \vspace{-10pt}
    
    \caption{\textit{Left:} Output space before (red) and after (green) the road-frame alignment. \textit{Right:} Truncation of visible lane boundary (blue) in a BEV perspective, only the green part is used for training.}
    \label{fig:modifications}
\end{figure}
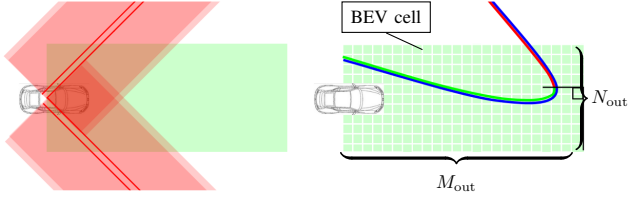

\subsection{3D Lane Detection}\label{sec:method:3dlane}
Due to its encouraging performance and low computational burden, we focus on the \bev detector proposed in~\cite{wang2023bev}. This method is capable of inference at rates of 185Hz (on a Tesla V100 GPU), and remains competitive with subsequent published works on 3D lane detection (c.f.,~\cite{pittner2024lanecpp,luo2023latr,bai2024curveformer}). It uses a ResNet backbone~\cite{he2016deep} to generate features that are passed through a Spatial Transformation Pyramid (STP)~\cite{wang2023bev}, akin to a Feature Pyramid Network~\cite{lin2017feature}. Different intermediary outputs in the STP are passed through a view relation module (VRM)~\cite{pan2020cross}, and are concatenated and convolved to produce four prediction heads (see Fig.~\ref{fig:networkarcitechture}). The input is a gray-scale image $I\in\Real^{W_{\mathrm{in}}\times H_{\mathrm{in}}}$, and the output is defined in a vehicle-fixed BEV perspective (see~\cref{fig:modifications}), comprising $ M_{\mathrm{out}}\times N_{\mathrm{out}}$ cells represented as tensors:
\begin{itemize}
    \item \textit{Confidence head}: $\Xbf^c\hspace{-2pt}\in\hspace{-2pt} [0,1]^{M_{\mathrm{out}}\hspace{-1pt}\times\hspace{-1pt} N_{\mathrm{out}}}$, likelihood that a cell in the BEV perspective contains a lane boundary.
    
    \item \textit{Embedding head}: $\Xbf^e\in \Real^{M_{\mathrm{out}}\times N_{\mathrm{out}} \times D_{\mathrm{out}}}$, a $D_{\mathrm{out}}$-dimensional embedding used for data association.
    
    \item \textit{Lateral offset head}: $\Xbf^o\in \Real^{M_{\mathrm{out}}\times N_{\mathrm{out}}}$, the proportion of a cell that is inside or outside the lane boundary.
    
    \item \textit{Elevation head}: $\Xbf^z\hspace{-2pt}\in\hspace{-2pt} \Real^{M_{\mathrm{out}}\hspace{-1pt}\times\hspace{-1pt} N_{\mathrm{out}}}$, a bias indicating the elevation of the lane in each cell of the BEV perspective.
\end{itemize}
During inference, a threshold on the confidence head determines the subset of cells that contain lane boundaries, the embedding head is used to cluster cells in this subset into unique lane boundaries, and a regression is done using the offset and elevation heads on cells associated with unique lane boundaries. This produces a parametric representation of the lane boundaries in the vehicle-fixed road frame $\{\mathcal{R}\}$.

\textbf{Losses and Training.}
The network is trained in a supervised manner with a loss comprising: a binary cross-entropy loss $\Lcal_{\text{conf}}^{\text{3D}}$ for the confidence head~\cite{wang2023bev}; an embedding loss $\Lcal_{\text{embed}}^{\text{3D}}$ that minimizes the mean distance of cell embeddings in the same cluster, and maximizes the variance of embeddings in different clusters analogous to~\cite{neven2018towards}; MSE losses in $\Lcal_{\text{offset}}^{\text{3D}}$ and $\Lcal_{\text{height}}^{\text{3D}}$; and a loss $\Lcal_{\text{lane}}^{\text{2D}}$ with segmentation and embedding losses in image space, identical to those proposed in~\cite{neven2018towards}. The total training loss is defined as in~\cite{wang2023bev}, with
\begin{subequations}\label{eq:lossfunction}
\begin{align}
\Lcal_{\text{total}} = &
\lambda_{\text{conf}}\Lcal_{\text{conf}}^{\text{3D}}+
\lambda_{\text{embed}}\Lcal_{\text{embed}}^{\text{3D}}+
\lambda_{\text{offset}}\Lcal_{\text{offset}}^{\text{3D}}+\\
&\lambda_{\text{height}}\Lcal_{\text{height}}^{\text{3D}}+
\lambda_{\text{lane}}\Lcal_{\text{lane}}^{\text{2D}},
\end{align}
\end{subequations}
where $\lambda_{i}>0$ are positive weights. We select the best model during training based on an F1-score defined as in~\cite{guo2020gen}, and use ResNet-34 as a nominal backbone, motivated by~\cite{wang2023bev}. %(refer to~\cref{app:training} for additional details).

\subsection{Adaptations for the Racing Application}\label{sec:method:adaptations}

In order to make existing methods such as BevLaneDet~\cite{wang2023bev} described in~\cref{sec:method:3dlane}, LATR~\cite{luo2023latr}, and PersFormer~\cite{chen2022persformer}, and AnchorLane~\cite{huang2023anchor3dlane} perform well on the \trilanes dataset, we need to make several modifications.

\textbf{Distortion and alignment.} The Lexus LC 500 test vehicle is instrumented with four Luxonis cameras with significant barrel distortion. As the camera intrinsics are known from external calibrations~\cite{furgale2013unified,rehder2016extending}, we undistort the images with a pre-calibrated fish-eye model~\cite{kannala2006generic} before passing them to the predictor (see conversion step in~\cref{fig:networkarcitechture}), permitting a single model to be used with multiple cameras.

Secondly, given that the cameras in the \trilanes dataset are not forward-facing, we align the BEV space with the road-frame, referred to as \emph{road-frame alignment} (see~\cref{fig:modifications}). This has two benefits: 1) it simplifies regression, as the predictions from any camera are defined in the same vehicle-fixed frame; 2) we generally find a larger number of cells with high confidence when aligning BEV perspective with the road frame, leading to improvements in recall (see~\cref{sec:adaptationsforracing}).

Finally, the visible lane boundaries sometimes curve back towards the car, breaking the interpolation on the ground truth data in the considered detectors~\cite{wang2023bev,chen2022persformer,luo2023latr,huang2023anchor3dlane}. We therefore augment the visibility to remove visible portions of the lane that curve back towards the car (see~\cref{fig:modifications}). 

\DeclareRobustCommand{\robustubar}[1]{\underaccent{\bar}{#1}}

\begin{figure}[t!]
    \centering
    \begin{tikzpicture}
        \node[inner sep=0pt] (image) at (0,0)
    {\includegraphics[width=\columnwidth, trim={0 1.6cm 0 1.6cm}, clip]{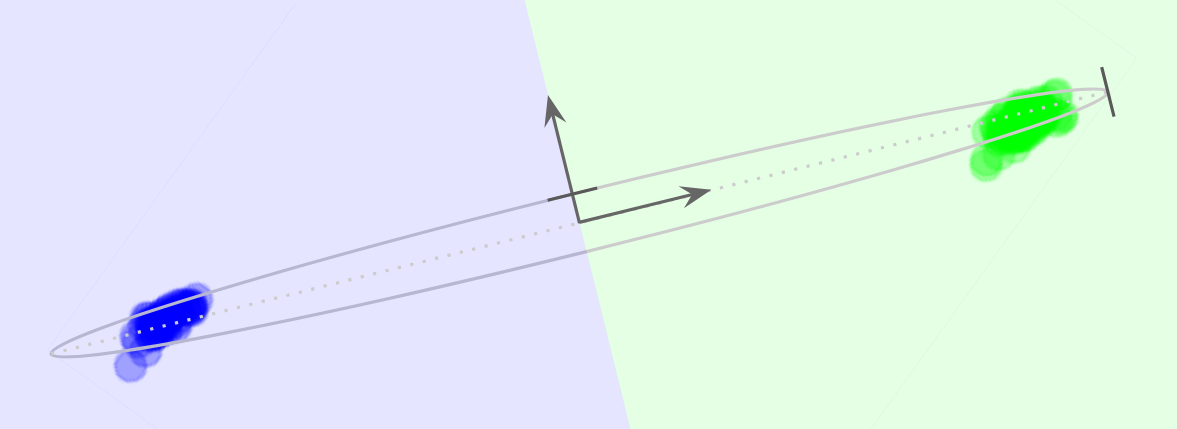}};
    \node[] at (1,0) {$\bar\vvec$};
    \node[] at (3.7,0.4) {$\bar\lambda$};
    \node[] at (0,0.8) {$\robustubar{\vvec}$};
    \node[] at (-0.5,0.2) {$\robustubar{\lambda}$};

    \node[circle,fill,inner sep=1.5pt, label=below:{\scalebox{.7}{$\mvec$}}] at (-0.06,-0.01){};
    \node[circle,fill,inner sep=2.5pt, green, label=right:{\scalebox{.7}{$\Xbf_{ij}^e$ when $c_{ij}=0$}}] at (-3.8,0.85){};
    \node[circle,fill,inner sep=2.5pt, blue, label=right:{\scalebox{.7}{$\Xbf_{ij}^e$ when $c_{ij}=1$}}] at (-3.8,0.55){};
    \end{tikzpicture}
    \vspace*{-13pt}
    
    \caption{A bimodal cluster in a $D_{\mathrm{out}}=2$-dimensional embedding head. Here, $\bar\lambda / \robustubar{\lambda}  \gg 1$ indicates that the prediction contains two lane boundaries. If detecting a single lane boundary, $\bar\lambda /\robustubar{\lambda} \approx 1$. }
    \label{fig:fig6}
\end{figure}
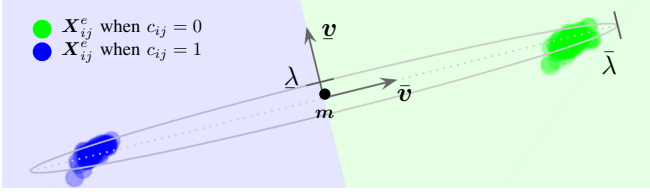

\textbf{Clustering.}
Clustering is an important step in the considered detectors, and a fast method analogous to a KNN~\cite{guo2003knn} was proposed for the \bev architecture in~\protect{\cite{wang2023bev}}.  We found this to be a computational bottleneck and used prior information that the track has at most two lane boundaries to speed up the clustering method. Specifically, we let 
\begin{equation}
\mathcal{X} =  \begin{Bmatrix}
\Xbf_{ij}^e\in\Real^{D_{\mathrm{out}}}\;:\;\Xbf_{ij}^e>\epsilon, \begin{matrix}
i=1,...,M_{\mathrm{out}}\\
j=1,...,N_{\mathrm{out}}
\end{matrix}\end{Bmatrix},
\end{equation}
be the set of features in the embedding head where the associated confidence is above a threshold $\epsilon > 0$. We compute the mean and covariance of the embeddings $\mvec=\tfrac{1}{|\mathcal{X}|}\sum_{\xvec\in\mathcal{X}}\xvec$ and $\C=\tfrac{1}{|\mathcal{X}|}\sum_{\xvec\in\mathcal{X}}(\xvec-\mvec)(\xvec-\mvec)^{\top}$, respectively. Let $(\bar\lambda, \bar\vvec)$ be the largest eigenvalue and corresponding eigenvector of $\C$, and let $(\ubar{\lambda},\ubar{\vvec})$ be its smallest eigenvalue/vector pair. We can then determine the number of clusters by threshold on $\bar\lambda/\ubar{\lambda}>\zeta>0$, and find to which cluster a cell belongs by a separating hyperplane with normal $\bar\vvec$ (see Fig.~\ref{fig:fig6}). Here,
\begin{equation}\label{eq:clusterlogic}
c_{ij} = \begin{cases}
0 & \bar\lambda \geq \zeta\ubar{\lambda} \;\land\; \bar\vvec^{\top}(\Xbf_{ij}^e - \mvec)>0,\\
1 & \text{otherwise}
\end{cases}
\end{equation}
indicates the cluster to which a BEV cell with embedding $\Xbf_{ij}^e$ belongs. The clustering for all cells in a prediction is $O(|\mathcal{X}|)$, resulting in significant speedups (see Sec.~\ref{sec:adaptationsforracing}).

\newcommand\nn{\hspace{4.5pt}}

\textbf{Model Optimization.}
Each component in Fig.~\ref{fig:networkarcitechture} needs to be optimized for vehicle deployments, to avoid frame drops and to minimize the utilization of computational resources. We therefore leverage TensorRT to compile and quantize the networks, with half-precision representations and post-training quantization (PTQ)~\cite{nagel2021white}. We also assess two different clustering algorithms: the method in~\cite{wang2023bev}, and our proposed PCA clustering, tailored for racing applications. As shown in~\cref{sec:adaptationsforracing}, the predictions (>$380$Hz) and a vectorized PCA clustering (>$1.2$kHz), can be run sequentially at rates exceeding 290Hz with negligible changes in key metrics. This exceeds the rates of detectors built purposely for speed, such as~\cite{qin2020ultra}, which only produce 2D detections.
\subsection{Ensemble Regression}\label{sec:method:regression}

Regression of parametric lanes to the predicted clusters is a key component in the considered 3D lane detectors~\cite{wang2023bev}, and usually this is done using low-order polynomial regression~\cite{guo2020gen, pittner2024lanecpp,chen2022persformer}. As we have access to multiple cameras, we propose a flexible approach that can be used for single cameras but that is also capable of fusing multiple independent predictions across cameras and time, leveraging IMU and odometry signals common in modern vehicles.  We refer to this methodology as \ourmethod (see~\cref{fig:networkarcitechture}).

\textbf{Representation.} Similar to~\cite{pittner2024lanecpp} we use Bezi\'er curves $\B:[0,1]\mapsto \Real^d$ as our lane boundary representation, with
\begin{equation}\label{eq:bezdef}
\B(\lambda; \pvec) =\sum_{i=0}^n
\begin{pmatrix}n\\i
\end{pmatrix}(1-\lambda)^{(n-i)}\lambda^i\bar\pvec_i,
\end{equation}
where $\pvec = (\bar\pvec_0;\cdots, \bar\pvec_n)\in\Real^{d(n+1)\times 1}$. However, unlike~\cite{pittner2024lanecpp}, we motivate this by two properties of the curve:
\begin{itemize}
    \item The Bezi\'er curve is {equivariant} under $\mathsf{SE(3)}$ when $d=3$ with respect to the control points $\pvec$.
    \item Regularizers such as the squared $\ell_2$-norm of a curvature is convex in the control points~\cite{de1978practical}.
\end{itemize}
The first point greatly simplifies interfacing the regressed outputs with downstream tasks, as we can transform the control points instead of the outputs evaluated at a specific $\lambda$. The second point is useful in that we can regularize the regression instead of including an additional loss during training, which was shown to have a marginal impact in~\cite{pittner2024lanecpp}. 

\textbf{Total Variation Regularization.}
To formalize the regression and introduce the regularizer, we consider a clustered prediction of a lane boundary in the road frame, as $\mathcal{D} = \{\yvec_i = (x_i,y_i,z_i)\in\Real^3\}_{i=1}^{|\mathcal{X}|}$, where $y_i$ and $z_i$ are computed from the lateral offset and elevation heads. We relate this to the domain of the curve by $\lambda_i = (x_i-x_i^-)(x_i^+-x_i^-)^{-1}$, where $x_i\in[x_i^-,x_i^+]$. We consider a simple linear model 
\begin{equation}\label{eq:regresionmodel}
\yvec_i = \B(\lambda_i; \pvec) + \epsbf_i,
\end{equation}
with Gaussian noise $\epsbf_i\sim\mathrm{N}(\Z,\Sigmabf_i)$, and define the loss
\begin{equation}\label{eq:regressionloss}
\begingroup
\color{white!70!black}
{\color{black}J(\pvec) \hspace{-2pt}=\hspace{-2pt} }\underbrace{{\color{black}\sum_{i=1}^{|\mathcal{X}|}\|\yvec_i \hspace{-1pt}-\hspace{-1pt} \B(\lambda_i; \pvec)\|_{\Sigmabf_i^{-1}}^2}}_{\text{negative log-likelihood}} \hspace{-1pt}{\color{black}+}\hspace{-2pt}
\underbrace{{\color{black}\sum_{k=1}^n \hspace{-1pt}\beta_k\hspace{-3pt}\int\hspace{-3pt}\|\B^{(k)}(\lambda; \pvec)\|_2^2\der\lambda}}_{\text{regularizer}},
\endgroup
\end{equation}
with $\beta_k\geq 0$. Implementing the regularizer with $\beta_2 >0$ and $\beta_k = 0$ for all $k\neq 2$ results in penalizing the total variation of the squared curvature, and we can similarly regularize higher-order path derivatives.~\cref{eq:regressionloss} is a small regularized weighted least squares problem. If the noise in~\eqref{eq:regresionmodel} is independent in the $d$ dimensions, this simplifies to solving $d$ smaller linear systems of size $n+1$.

\textbf{Regression over Time and Cameras.} Given that past predictions of the road geometry may inform present inference, we relate the multiple predictions by gyroscopic measurements $\omegabf(t)\in\Real^3$ from an IMU and the vehicle velocity measurements $\vvec(t)\in\Real^3$ in $\{\mathcal{R}\}$ computed from the wheel speeds. These measurements are available in the \trilanes dataset at 500Hz and 62.5Hz, respectively. For short intervals $t\in[t_{k},t_{k+1}]$ of length $\Delta_k=t_{k+1}-t_{k}$, we let $\vvec_k=\vvec(t_k)$ and $\omegabf_k=\omegabf(t_k)$ and integrate a transform
\begin{equation}\label{eq:matrixexponential}
\T_k=\mathrm{exp}\begin{pmatrix}
\begin{bmatrix}
[\Delta_k\omegabf_{k}]^{\land}_{\mathsf{SO(3)}} & \Delta_k\vvec_{k} \\
\Z & 1
\end{bmatrix}^{\land}_{\mathsf{SE(3)}}
\end{pmatrix}
\in\mathsf{SE(3)},
\end{equation}
that relates the road-frames at the end-points of the time interval, where $[\cdot]^{\land}_{G}$ is the {hat operator}, spanning the Lie algebra associated with the Lie group $G$~\cite{murray2017mathematical}. In practice, we compute~\eqref{eq:matrixexponential} analytically, or with an expansion in $\eta=\|\Delta_k\omegabf_k\|$ for small $\eta$, and maintain a short buffer of such transforms, integrating them as new measurements come in. This is related to the pre-integration techniques commonly employed in visual-inertial odometry~\cite{forster2015imu,forster2016manifold}. We use this buffer to transform the most recent clustered lane predictions to the same current road frame. In the following, $\mathcal{D}_{k|s}$ is the clustered output at $t_{s}$ and is transformed into the road frame at time $t_k$. To perform a regression over the $N_{\mathrm{buff}}$ most recent predictions computed for images sampled at $t_{s_i}$ with $i=1,...,N_{\mathrm{buff}}$, we minimize~\eqref{eq:regressionloss} with respect to $\bar\Dcal = \bigcup_{i=1}^{N_{\mathrm{buff}}}\mathcal{D}_{k|s_i}$. Furthermore, as the individual predictions from various cameras are done in the same (moving) road frame, we can align predictions from different cameras in the same way and regress over both cameras and time (see~\cref{fig:networkarcitechture}).

\iffalse
\begin{subequations}
\textbf{Quantifying Uncertainty.} The minimizer of~\eqref{eq:regressionloss} can be written as a linear map from $\yvec = (\yvec_1; ...; \yvec_m)$ to $\hat\pvec = \text{argmin}_{\pvec} J(\pvec)$. Specifically, there are matrices $\Q\succeq \Z$ and $\{\A_i\}_{i=1}^m$ of appropriate dimensions such that
\begin{align}
\hat\pvec \hspace{-1pt}&=\hspace{-1pt} \Big(
    \Q + \sum_{i=1}^{|\bar{\mathcal{D}}|}\A_i^{\top}\Sigmabf_i^{-1}\A_i
\Big)^{-1}
 \Big(
\sum_{i=1}^{|\bar{\mathcal{D}}|}\A_i^{\top}\Sigmabf_i^{-1}\yvec_i
 \Big)\hspace{-1pt}.\hspace{-1pt}
\end{align}
\end{subequations}
Additionally, under the assumption that there is a true Bezi\'er curve representing a lane boundary with parameters $\pvec$, and that $\hat\pvec\hat\pvec^{\top}\approx \pvec\pvec^{\top}$ (when $\beta_i>0$), then there are matrices $\M$ and $\N_i$, such that the covariance of the error $\tilde\pvec=\pvec - \hat\pvec$ is
\begin{equation}\label{eq:covestimate}
\text{Cov}(\tilde\pvec, \tilde\pvec) \approx \M\hat\pvec\hat\pvec^{\top}\M^{\top} + \sum_{i=1}^{|\bar{\mathcal{D}}|}\N_i\hat{\Sigmabf}_i\N_i^{\top}\triangleq \C(\hat\pvec),
\end{equation}
where $\hat\Sigmabf_i$ is an estimate of the covariance of $\epsbf_i$, which can be approximated from the many overlapping ensemble predictions in $\bar{\mathcal{D}}$. %A method for doing so is derived in~\cref{app:curvature}, but if assumed constant, this covariance matrix may also be tuned by hand with respect to consistency metrics of interest. The latter comes with no additional computational overhead at inference time.
\fi 
\section{Experiments}\label{sec:results}

To test the proposed \ourmethod methodology, we first conduct experiments with various existing DL methods for 3D lane detection on the \trilanes dataset, focusing on classification scores, near-vehicle predictions, and compute times. We appropriate the metrics in~\cite{guo2020gen} (also used in~\cite{chen2022persformer,wang2023bev,pittner2024lanecpp}). We start by adapting \bev, LATR, PersFormer, and AnchorLane detectors for racing and study how the proposed modifications affect performance in~\cref{sec:adaptationsforracing}. Next, we assess the impact of the multi-camera regression in~\cref{sec:regresson:ablations} in a set of ablation studies where the method is deployed in a vehicle on track (the hold-out split).% (see~\cref{app:datasplits}). 

Inference is done with a C++ implementation of the pipeline in ROS2 using a single RTX 3070 (or 4070 Ti S) GPU and an Intel Xeon CPU (see ~\cref{fig:networkarcitechture}). The in-vehicle experiments are done on the Lexus LC 500 with an expert driver maneuvering the vehicle at pace (see~\cref{fig:flagship}.a). When done across cameras and/or time, the ground truth is a union of the visible lane boundaries across cameras and/or time.

\begin{table}[t!]
    \centering
    %\resultQ{Table Showing performance with Persformer, LATR, Bev-Lane-Det, distorted images from thunderhill.}
    \caption{\textbf{Performance on the \trilanes dataset:} The $^{\star}$ indicates that the modification in~\cref{sec:method:adaptations} are applied to the model. 
    }
    \vspace{-10pt}
    
    \label{tab:testTHdata}
        \setlength{\tabcolsep}{5.5pt}
        \resizebox{\columnwidth}{!}{\begin{NiceTabular}{@{}cccccccc}[
            code-before =%
            \rectanglecolor{\bestcolor}{5-2}{5-2}
            \rectanglecolor{\bestcolor}{4-3}{4-3}
            \rectanglecolor{\bestcolor}{5-4}{5-6}
            \rectanglecolor{\bestcolor}{3-7}{3-7}
            \rectanglecolor{\bestcolor}{2-7}{2-7}
            \rectanglecolor{\bestcolor}{2-8}{2-8}
            \rectanglecolor{\secondbestcolor}{7-2}{7-2}
            \rectanglecolor{\secondbestcolor}{2-5}{2-5}
            \rectanglecolor{\secondbestcolor}{3-5}{3-5}
            \rectanglecolor{\secondbestcolor}{5-3}{5-3}
            \rectanglecolor{\secondbestcolor}{3-4}{3-4}
            \rectanglecolor{\secondbestcolor}{2-6}{2-6}
            \rectanglecolor{\secondbestcolor}{3-8}{3-8}
        ]
        \toprule
        {\header{Method}} & {\header{F1} $\uparrow$} & {\header{AP} $\uparrow$} & {\header{AR} $\uparrow$} & {\header{Y (near)}} & {\header{Y (far)}} & {\header{Z (near)}} & {\header{Z (far)}} \\ \midrule
        %\header{PersFormer}~\cite{chen2022persformer}  & 8.18 & -- & -- & -- & -- & -- & -- \\
        \header{LATR}~\cite{luo2023latr}  & 78.63 & 75.08 & 82.52 & 0.16 & 0.56 & 0.06 & 0.42 \\
        \header{LATR}$^{\star}$  & 85.65 & 79.81 & 92.40 & 0.16 & 0.57 & 0.06 & 0.44 \\
        \header{BevLaneDet}~\cite{wang2023bev} & 69.64 & 98.64 & 53.81 & 0.22 & 1.35 & 0.16 & 1.32 \\
        \header{BevLaneDet$^\star$} &  91.28 & 89.15 &  93.50 &  0.14 & 0.53 &  0.09 & 0.44 \\
        \header{AnchorLane}~\cite{huang2023anchor3dlane} & 60.30 & 71.60 & 52.10 & 0.23 & 1.41 & 0.11 & 1.29 \\
        \header{AnchorLane}$^{\star}$ & 83.99 & 86.39 & 72.20 & 0.19 & 0.62 & 0.08 & 0.47 \\
        \bottomrule
    \end{NiceTabular}}
\end{table}

\newcommand\persformer{\textsc{PersFormer}\xspace}
\newcommand\anchorlane{\textsc{AnchorLane}\xspace}
\newcommand\latr{\textsc{LATR}\xspace}

\subsection{Modifications for Racing}\label{sec:adaptationsforracing}

We start by motivating the use of \bev for the racing application and study the impact of the modifications in~\cref{sec:method:adaptations}. The evaluations are done on single monocular predictions without the regression in~\cref{sec:method:regression}, with the metrics in~\cite{guo2020gen} computed on the test split of \trilanes. For reference, we compare the results with the anchor-based \persformer~\cite{chen2022persformer}, the anchor-based \anchorlane~\cite{huang2023anchor3dlane}, and the anchor-free \latr~\cite{luo2023latr} detectors. We did not consider the more recent methods~\cite{pittner2024lanecpp,pittner2025sparselanestp}, as code was not available.

\textbf{Racing Modifications.} We evaluate the considered methods on the subset of \trilanes associated with camera 0, with results in~\cref{tab:testTHdata}. \persformer~\cite{chen2022persformer} was tested but never converged during training on \trilanes (F1 score never exceeded 10\%), likely due to its reliance on anchors tailored for regular driving. \latr performs much better, as it learns the camera extrinsics and does not have the surrogate representations of \persformer. Its performance improves when implementing the modifications in~\cref{sec:method:adaptations}. With the nominal \bev, precision is high, but the low recall impacts the overall F1 score. Our modified variant of \bev recovers recall, with performance on par with the method trained on synthetic Apollo data~\cite{huang2020apollo} (c.f.,~\cite{wang2023bev}).

\textbf{Ensemble Predictions.} With the encouraging performance of the modified \bev on \trilanes, we consider if it is feasible to train a single detector operating on all the image feeds, or if different models should be trained for specific cameras, inferring the lanes in an ensemble fashion. To assess this, we train networks for each camera individually, for each side-facing pair of cameras, and for all cameras simultaneously%(see~\cref{app:ensemble:training} for details)
. The results are reported in~\cref{tab:multicamerapredictions}, and we observe marginally worse performance when a model is trained on images from many different cameras. This is likely due to the virtual camera, a module that unifies the various cameras by ``averaging'' camera parameters~\cite{wang2023bev}. Based on this, we elect to train separate models for each camera.

\begin{table}[t!]
 \vspace{-5pt}
    % \vspace{-5pt}
    \centering
    %\resultQ{Impact of changes and camera-specific networks}
    \caption{\textbf{Training on single or multiple camera feeds}. }
    \vspace{-10pt}
    
    \label{tab:multicamerapredictions}
        \setlength{\tabcolsep}{5.5pt}
        \resizebox{\columnwidth}{!}{\begin{NiceTabular}{@{}cccccccc}[
            code-before =%
            \rectanglecolor{\bestcolor}{4-2}{4-4}%
            \rectanglecolor{\bestcolor}{3-5}{3-5}%
            \rectanglecolor{\bestcolor}{5-6}{5-6}%
            \rectanglecolor{\bestcolor}{3-7}{3-7}%
            \rectanglecolor{\bestcolor}{2-8}{2-8}%
            \rectanglecolor{\bestcolor}{5-8}{5-8}%
            \rectanglecolor{\secondbestcolor}{5-2}{5-4}%
            \rectanglecolor{\secondbestcolor}{2-5}{2-5}%
            \rectanglecolor{\secondbestcolor}{6-5}{6-5}%
            \rectanglecolor{\secondbestcolor}{4-6}{4-6}%
            \rectanglecolor{\secondbestcolor}{2-6}{2-6}%
            \rectanglecolor{\secondbestcolor}{2-7}{2-7}%
            \rectanglecolor{\secondbestcolor}{6-7}{6-7}%
        ]
        \toprule
        {\header{Cam}} & {\header{F1} $\uparrow$} & {\header{AP} $\uparrow$} & {\header{AR} $\uparrow$} & {\header{Y (near)}} & {\header{Y (far)}} & {\header{Z (near)}} & {\header{Z (far)}} \\ \midrule
        0 & 91.28 & 89.15 &  93.50 &  0.14 & 0.53 &  0.09 & 0.44 \\  
        1 & 92.32 & 90.00 &  94.77 & 0.13 & 0.54 &   0.08 & 0.46 \\
        2 & 95.19 &  95.45 & 94.93 &  0.15 & 0.53 & 0.10 & 0.45 \\
        3 & 95.08 & 95.31 & 94.85 & 0.15 & 0.52 &  0.10 &   0.44 \\
        $\{0,1\}$ & 91.61 & 89.73 & 93.57 & 0.14 & 0.54 & 0.09 &  0.45 \\
        $\{2,3\}$ & 94.80 & 95.18 & 94.42 & 0.15 &  0.54 & 0.10 & 0.45 \\
        $\{0,1,2,3\}$ &  92.35 & 91.61 &  93.08 & 0.17 & 0.57 & 0.10 &  0.45 \\
        \bottomrule
    \end{NiceTabular}}
\end{table}

\begin{table}[t!]
 \vspace{-5pt}
    % \vspace{-5pt}
    \centering
    %\resultQ{Showing performance across methods + optimization techniques}
    \caption{\textbf{Turnaround rates:} HP -- Half Precision, PTQ -- Post Training Quantization, PCA -- Use the PCA clustering method.}
    \vspace{-10pt}
    
    \label{tab:optimization}
    \setlength{\tabcolsep}{5.5pt}
    \resizebox{\columnwidth}{!}{\begin{NiceTabular}{@{}ccc|cc|c|cc}[
    code-before =%
            %\rectanglecolor{gray!20}{4-1}{4-13}%
            %\rectanglecolor{gray!20}{6-1}{6-13}%
            %\rectanglecolor{gray!20}{8-1}{8-13}
            \rectanglecolor{\bestcolor}{8-4}{8-6}%
            \rectanglecolor{\bestcolor}{7-7}{7-7}%
            \rectanglecolor{\bestcolor}{3-8}{8-8}%
            \rectanglecolor{\secondbestcolor}{7-4}{7-5}%
            \rectanglecolor{\secondbestcolor}{4-6}{4-6}%
            \rectanglecolor{\secondbestcolor}{8-7}{8-7}%
            \rectanglecolor{\secondbestcolor}{5-7}{5-7}%
            \rectanglecolor{\secondbestcolor}{3-7}{3-7}%
    ]
        \toprule
        \multicolumn{3}{c}{\header{Optimization}} &  \multicolumn{2}{c}{\header{Prediction (Hz)}} & \header{Clustering (Hz)} & \multicolumn{2}{c}{\header{Metrics}}\\
        \header{HP}  &
        \header{PTQ} & 
        \header{PCA} & \header{RTX 3070} & \header{RTX 4070} & \header{CPU} & \header{F1} $(\uparrow)$  & \header{Y (near)}\\
        \midrule
    \xmark & \xmark & \xmark & \nn 67.7  & 129.4  & \nn\nn 59.2 & 93.48 & 0.14 \\
    \xmark & \xmark & \cmark & \nn 73.6  & 132.2 & 1221.2 & 93.41 & 0.14 \\
    \cmark & \xmark & \xmark& 103.0 &  189.4 & \nn\nn 59.7 & 93.48 & 0.14 \\
    \cmark & \xmark & \cmark & 106.3 &  190.3 & 1217.3 & 93.41 & 0.14 \\
    \cmark & \cmark & \xmark & 251.7 & 379.0 & \nn\nn 61.0 & 93.49 & 0.14 \\
    \cmark & \cmark & \cmark & 253.9 & 383.9 & 1234.4  & 93.48 & 0.14 \\
    \bottomrule
    \end{NiceTabular}}
\end{table}

\textbf{Model Optimization.} We consider various model optimizations, including PTQ through TensorRT~\cite{nagel2021white} and changing the clustering algorithm to the PCA method proposed in~\cref{sec:adaptationsforracing}. The turnaround rates for the components of Fig.~\ref{fig:networkarcitechture} are computed in the vehicle for six consecutive laps on the racetrack from a vehicle deployment in~\cref{tab:optimization}. The metrics are otherwise computed on the test split of \trilanes. % (see~\cref{app:modeloptimization} for additional details).

There is a small performance decrease in the ``far'' MAEs with PCA clustering compared to the clustering in~\cite{wang2023bev}, but no significant change in the ``near'' MAEs or the F1 score. A small price for the large speed-up in clustering (>20x). The metrics do not change significantly with half-precision or PTQ, but we get a speed-up (>3x) on the 3070 and 4070 GPUs. Consequently, we use a half-precision optimized model with PCA clustering when analyzing the ensemble method.
 
%\resultQ{Table Showing performance across various backbones + optimization techniques}

\subsection{Ensemble Regression and Hardware Deployment}\label{sec:regresson:ablations}

In these experiments, we report results on the hold-out split: whole laps that have not been seen during training, with different calibrations, sensor placements, and lighting conditions. Our nominal \ourmethod method uses: (i) one modified \bev network trained per camera, optimized with half-precision and PTQ; (ii) the proposed PCA clustering method; and (iii) regression with a buffer of length $N_{\mathrm{buff}}=8$, 5th order Bezi\'er curves, and a TV regularizer in~\cref{eq:regressionloss} with $\beta_3=\beta_4=10^{-3}$ and $\beta_i=0$ otherwise.

\begin{table}[t!]
    \centering
    \caption{\textbf{Optimization Ablations (single-camera predictions)}
    }
    \vspace{-10pt}
    
    \label{tab:regressionoptimization:single}
    \setlength{\tabcolsep}{5.5pt}
    \resizebox{\columnwidth}{!}{\begin{NiceTabular}{@{}ccc|ccccccc}[
    code-before =%
            \rectanglecolor{\bestcolor}{6-4}{6-8}%
            \rectanglecolor{\bestcolor}{3-9}{8-9}%
            \rectanglecolor{\bestcolor}{6-10}{6-10}%
            \rectanglecolor{\bestcolor}{8-10}{8-10}%
            \rectanglecolor{\secondbestcolor}{5-4}{5-8}%
    ]
        \toprule
        \multicolumn{3}{c}{\header{Optimization}} & \multicolumn{7}{c}{\header{Metrics}}\\
        \header{HP}  &
        \header{PTQ} & 
        \header{PCA} & \header{F1} $(\uparrow)$ & \header{AP} $(\uparrow)$  & \header{AR} $(\uparrow)$ & \header{Y (near)} & \header{Y (far)} & \header{Z (near)} & \header{Z (far)} \\
        \midrule
    \xmark & \xmark & \xmark & 84.93 &       83.46 &    86.46 &            0.27 &          0.80 &            0.20 &          0.64 \\
    \xmark & \xmark & \cmark & 85.23 &       83.60 &    86.92 &            0.28 &          0.80 &            0.20 &          0.63 \\
    \cmark & \xmark & \xmark & 87.40 &       85.59 &    89.29 &            0.26 &          0.78 &            0.20 &          0.65 \\
    \cmark & \xmark & \cmark & 87.48 &       85.74 &    89.30 &            0.26 &          0.76 &            0.20 &          0.62 \\
    \cmark & \cmark & \xmark &  85.11 &       83.68 &    86.59 &            0.28 &          0.82 &            0.20 &          0.66 \\
    \cmark & \cmark & \cmark &  85.09 &       83.43 &    86.81 &            0.28 &          0.79 &            0.20 &          0.62 \\
    \bottomrule
    \end{NiceTabular}}
\end{table}

\begin{table}[t!]
    \vspace{-5pt}
    \centering
    \caption{\textbf{Optimization Ablations (ensemble predictions)}
    }
    \vspace{-10pt}
    
    \label{tab:regressionoptimization}
    \setlength{\tabcolsep}{5.5pt}
    \resizebox{\columnwidth}{!}{\begin{NiceTabular}{@{}ccc|ccccccc}[
    code-before =%
            \rectanglecolor{\bestcolor}{6-4}{6-8}%
            \rectanglecolor{\bestcolor}{5-7}{5-8}%
            \rectanglecolor{\bestcolor}{4-9}{4-9}%
            \rectanglecolor{\bestcolor}{6-9}{6-9}%
            \rectanglecolor{\bestcolor}{8-9}{8-9}%
            \rectanglecolor{\bestcolor}{3-10}{6-10}%
            \rectanglecolor{\bestcolor}{8-10}{8-10}%
            \rectanglecolor{\secondbestcolor}{5-4}{5-8}%
    ]
        \toprule
        \multicolumn{3}{c}{\header{Optimization}} & \multicolumn{7}{c}{\header{Metrics}}\\
        \header{HP}  &
        \header{PTQ} & 
        \header{PCA} & \header{F1} $(\uparrow)$ & \header{AP} $(\uparrow)$  & \header{AR} $(\uparrow)$ & \header{Y (near)} & \header{Y (far)} & \header{Z (near)} & \header{Z (far)} \\
        \midrule
    \xmark & \xmark & \xmark & 87.59 &       80.35 &    96.27 &            0.19 &          0.73 &            0.22 &          0.65 \\
    \xmark & \xmark & \cmark & 88.97 &       82.32 &    96.80 &            0.19 &          0.74 &            0.20 &          0.65 \\
    \cmark & \xmark & \xmark & 90.28 &       83.11 &    98.80 &            {0.17} &          0.69 &            0.21 &          0.65 \\
    \cmark & \xmark & \cmark & {90.90} &       84.08 &    98.94 &            {0.17} &          0.69 &            0.20 &          0.65 \\
    \cmark & \cmark & \xmark & 87.29 &       79.81 &    96.31 &            0.19 &          0.74 &            0.21 &          0.66 \\
    \cmark & \cmark & \cmark & 89.25 &       82.91 &    96.64 &            0.18 &          0.73 &            0.20 &          0.65 \\
    \bottomrule
    \end{NiceTabular}}
\end{table}

\iffalse
\placeholderfig{
Qualitative example showing single vs multi-cam preds
}{fig:qualitativeregression}
{Qualitative example showing individual predictions (red, corresponding with~\cref{tab:regressionoptimization:single}) and the regressed output (green, corresponding with~\cref{tab:regressionoptimization}) over the ground truth (blue).}
{0.6}
{0.3}
\fi

\newcommand{\hwplotA}{\raisebox{2pt}{\tikz{\draw[red,solid,line width=1.5pt,opacity=0.3](0,0) -- (5mm,0);}}}
\newcommand{\hwplotB}{\raisebox{2pt}{\tikz{\draw[green,dash dot,line width=1.5pt](0,0) -- (5mm,0);}}}
\newcommand{\hwplotC}{\raisebox{2pt}{\tikz{\draw[blue,line width=2.0pt](0,0) -- (5mm,0);}}}

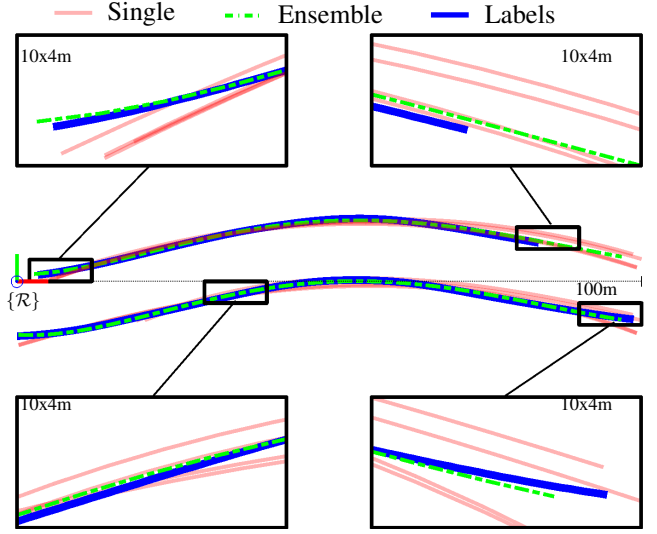
\begin{figure}
    \centering
    \begin{tikzpicture}
    \node[anchor=west] at (0,0) {\includegraphics[width=\columnwidth, clip, trim={1.8cm 1.2cm 1cm 0}]{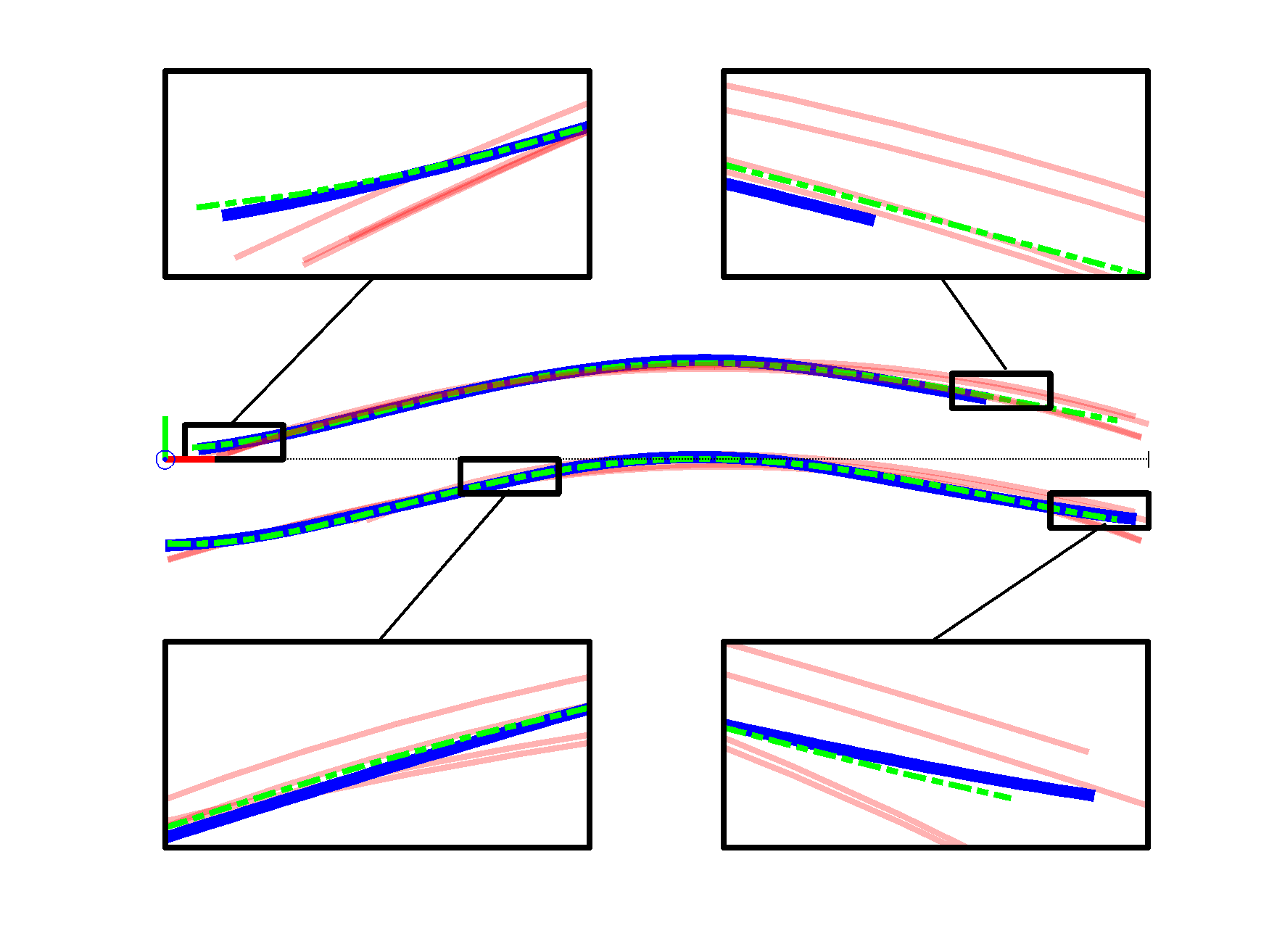}};
    \node[anchor=north] at (0.3,-0.3) {\scalebox{.7}{$\{\mathcal{R}\}$}};
    \node[anchor=north] at (7.88,-0.18) {\scalebox{.7}{100m}};
    \node[anchor=north] at (0.6,2.9) {\scalebox{.7}{10x4m}};
    \node[anchor=north] at (7.75,2.9) {\scalebox{.7}{10x4m}};
    \node[anchor=north] at (0.6,-1.7) {\scalebox{.7}{10x4m}};
    \node[anchor=north] at (7.75,-1.7) {\scalebox{.7}{10x4m}};
    \node[] at (1.5,3.25) {\hwplotA\hspace{5pt}Single};
    \node[] at (4,3.25) {\hwplotB\hspace{5pt}Ensemble};
    \node[] at (6.5,3.25) {\hwplotC\hspace{5pt}Labels};
    \end{tikzpicture}
    \vspace{-20pt}
    
    \caption{BEV predictions (red, corresponding with~\cref{tab:regressionoptimization:single}) and the regressed output (green, corresponding with~\cref{tab:regressionoptimization}) over the lane label (blue).}
    \label{fig:qualitativeregression}
\end{figure}

The results with individual camera and ensemble predictions are reported in~\cref{tab:regressionoptimization:single} and~\cref{tab:regressionoptimization}, respectively. In both cases, the half-precision model is more amenable to regression than the PTQ model, and the PCA clustering generally improves the classification metrics. We speculate that this is due to the effective time span of the cluster buffer increasing as processing time increases. Numerically integrating the odometry results in local truncation errors that accumulate over time, affecting the metrics. The regression improves performance over individual predictions (c.f.~\cref{tab:regressionoptimization:single} and~\cref{tab:regressionoptimization}) by $\sim$3 points in the F1 measure, and $>30\%$ reduction in the Y-near MAE consistently across the 6 sets of models, an example of this performance gap is visualized in \cref{fig:qualitativeregression}. The multi-camera regression has an averaging effect, partly explaining the improvements in~\cref{tab:regressionoptimization}.

\subsection{Limitations}\label{sec:limitatoins}
A limitation of this work is that we consider a single 2-mile racetrack. This is due to the fundamental lack of real racing datasets with both 3D lane labels and IMU data. To address this, we evaluated the method on the {hold-out split} (12 miles of driving data collected at a separate occation to the training/test data). We noted a small performance decrease (c.f.,~\cref{tab:optimization} and~\cref{tab:regressionoptimization:single}). Despite this limitation, it is clear that using IMU and multiple camera feeds improves 3D lane predictions (c.f.,~\cref{tab:regressionoptimization:single} and~\cref{tab:regressionoptimization}). Another limitation is the lack of camera synchronization and absence of LiDAR data, which makes it difficult to compare to methods such as PETRv2~\cite{liu2023petrv2}. A general drawback of the \ourmethod method is the need for inertial measurements, which are required to relate the clusters associated with multiple cameras (if they are not synchronized) or when regressing over time. Finally, the method assumes a static environment, in which no time-varying or otherwise dynamic observations are present. %These signals are common in modern vehicles, but may be cumbersome to splice out from the relevant buses, and are not present in most datasets for 3D lane detection~\cite{huang2020apollo,yan2022once}.

\section{Conclusions}\label{sec:conclusion}

We proposed the ensemble 3D lane detection method \ourmethod trained on the new \trilanes dataset. Using this approach, we can run inference sequentially at rates exceeding 290Hz on a single GPU. We achieve F1 scores  $>0.9$ on the hold-out split of \trilanes, which can be compared with the typical performance of lane detection on real data with F1 scores of 0.5-0.7 (c.f.~\cite{wang2023bev}). Our high classification scores may be due to lesser variance in the training data distribution. Another reason may be that the labels generated by the automatic labeling pipeline are of high quality, as they are mapped with cm-level accuracy using the OxTS and thus perhaps more conducive to training. We noted strict improvements in performance when using IMU and vehicle speed measurements (c.f.,~\cref{tab:regressionoptimization:single} and~\cref{tab:regressionoptimization}) in the multi-camera regression, and speculate that \ourmethod can improve lane predictions in regular AD driving. Assessing this requires additional data or augmentation of existing AD datasets, and will be the subject of future work.

\bibliography{references}{}
\bibliographystyle{IEEEtran}

\appendix
%\iffalse

% \setcounter{page}{1}
% \maketitlesupplementary

\section{The \trilanes Dataset}\label{app:dataset}

\trilanes consists of a set of ``runs'', each corresponding to one or two laps on the racetrack. To generate a consistent temporal alignment of the camera and IMU data\footnote{The Camera and IMU are logged directly in ROS2 on a 12th Gen Intel I7-1270P (base frequency 2.2GHz) CPU Linux laptop.} and ground truth data from the OxTS.\footnote{The CAN data and vehicle pose from the OxTS are logged through a dSpace MicroAutoboxII (DS1401) and published to ROS2 running on an Intel Xeon E-2278GE (base frequency 3.30GHz) CPU Linux computer.} We use an NTP timeserver with a common clock provided using a TM2000 device. The IMU and cameras are detachable from the vehicle, and their exact location in the vehicle-fixed frame may vary between runs. To account for this, the camera intrinsics and camera-IMU extrinsics are computed before each run. The IMU is used to align the sensor frame $\{\Scal\}$ with the road frame $\{\mathcal{R}\}$ by physically adjusting the sensors on the car. As such, the transform $\{\Scal\}\to \{\mathcal{R}\}$ is known down to a small rotation about the $z$-axis. This slight bias is estimated and removed, usually corresponding to a rotation of a few degrees. Within the datasets, we noted that the elevation estimate from the OxTS was biased by up to 30cm, depending on the calibration. As such, for each run, we tune away this bias in the $\{\Gcal\}\to\{\mathcal{R}\}$ transform based on an alignment of the projected 3D ground truth lanes into distorted image space. Thus, we know the transform chain $
   \{\Gcal\}\to\{\mathcal{R}\}\to\{\Scal\}\to\{\Ccal_0\},\{\Ccal_1\},\{\Ccal_2\},\{\Ccal_3\}$,
which is used for the automatic labeling. In this process, the high-precision map in $\{\Gcal\}$ is projected into the camera frames $\{\Ccal_i\}_{i\in\{0,1,2,3\}}$ to determine the lane boundary visibility. The visible subset of the lane boundaries is then stored in $\{\mathcal{R}\}$ to conform with existing dataset formats~\cite{huang2020apollo,yan2022once}. A 2D projection of the racetrack is shown in~\cref{fig:racetrack}, illustrating the lane labels in distorted image space, showing the field of view of Camera $1$, and the high quality of the labels. A distance of $d_{\max}=80$m is chosen as the lane labels approach the vanishing point in image space at this distance.

\def\widthA{4.9}
\def\widthB{3.5}
\def\imageheightsep{2.37}
\def\imageheightoffset{0.6}

\newcommand\includemapcamfigure[3]{
\node[inner sep=0pt, anchor=south west] (#1) at (\widthA, {\imageheightoffset+#2*\imageheightsep})
    {\includegraphics[width=\widthB cm,  clip, trim={2cm, 1.8cm, 2cm, 0}]{#3}};
\node[anchor=south] at ({\widthA + 0.5*\widthB}, {\imageheightoffset+#2*\imageheightsep+\imageheightsep*0.65}) {{\scriptsize cam. #2}};
}
\begin{figure}[t!]
\centering
\iffalse
\begin{tikzpicture}

\node[inner sep=0pt, anchor=south west] (map) at (0,0)
    {\includegraphics[width=\widthA cm, clip, trim={0, 2cm, 1cm, 0}]{figures/2D_map_view.png}};

\includemapcamfigure{cA}{0}{figures/camera_perspective_1.png}
\includemapcamfigure{cB}{1}{figures/camera_perspective_2.png}
\includemapcamfigure{cC}{2}{figures/camera_perspective_3.png}
\includemapcamfigure{cD}{3}{figures/camera_perspective_4.png}

\draw[black, thick] ({\widthA+0*\widthB}, {\imageheightoffset+1*\imageheightsep}) rectangle ({\widthA+\widthB}, {\imageheightoffset+1.96*\imageheightsep});
\draw[black, dotted] (3.2,5.0) -- ({\widthA+0*\widthB},  {\imageheightoffset+1*\imageheightsep+\imageheightsep*0.5});
\end{tikzpicture}
\fi 
\includegraphics[width=\columnwidth]{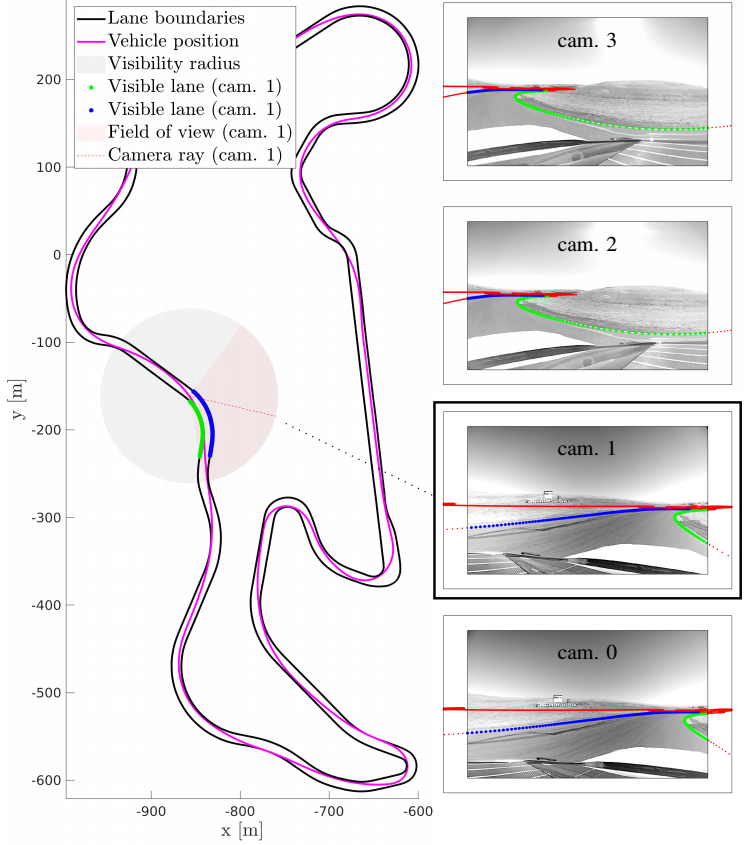}
\vspace{-20pt}

\caption{Visualization of the racetrack and labels. \textit{Left:} Mapped lane boundaries (black) with the vehicle position from one run in the dataset (magenta). The visibility radius (gray) limits the portion of the outer (green) and inner (blue) lane boundaries that are visible in each camera. The field of view of camera 1 is shown in red. \textit{Right}: Images with the outer (green) and inner (blue) lane boundaries in distorted image space. The lane boundaries that are not visible in each camera image are shown in red.}
\label{fig:racetrack}
\end{figure}

\textbf{Inconsistencies.} There are a few ambiguous points along the track where the lane boundary is not well-defined. Such as the outer lane around $(-600,-350)$. In this region, we define the lane boundary as a smooth interpolation between the end of one marked boundary and the beginning of the next. Furthermore, there are sharp curves along the track where the visible lane boundary is not an injective function along the principal ray of the camera. Here, there will be a discontinuous jump in the lane labels as a function of time, which will affect the local performance of the 3D lane predictor, as there will be consecutive images that ``disagree'' on where the lane boundary ends. Such inconsistencies are present in $\lesssim 0.1\%$ of the images, occurring more frequently in cameras 0 and 1. This is one explanation for the difference in performance between cameras $\{0,1\}$ and cameras \{2,3\}.

\textbf{Data Format.} The automatic labeling procedure is then done as described in Sec.~\ref{sec:data}, and the data is organized by runs. Some of these runs constitute the test/train split, and other runs are used to evaluate in-vehicle deployments and predictor generalization (see~\cref{tab:datasplits}). Lane annotations are stored in the format of ApolloScape~\cite{huang2020apollo} to facilitate integration and testing with existing methods. Additionally, we annotate each image with a time-stamped $\mathsf{SE(3)}$ transforms $\T_{\Gcal\to\mathcal{R}}$, $\T_{\mathcal{R}\to\mathcal{C}_i}$, $\T_{\Gcal\to\mathcal{C}_i}$, measured with an OxTS GNSS receiver~\cite{oxts2024} and computed using the offline calibrations. These transforms are interpolated in the time stamps of the cameras, which are sampled asynchronously at 20Hz. We also provide IMU measurements in the sensor frame $\{\mathcal{S}\}$ at 500Hz. As such, the \trilanes dataset has value both as a benchmark in 3D lane detection and may also be useful in exploring visual-inertial simultaneous localization and mapping in a racing setting.

\begin{table}[h!]
    \centering
    \caption{\textbf{Data splits and partitioning :} The various runs constituting \trilanes, indicating which runs were used for evaluations, and which subset of the train/test data was used to train the models evaluated in the October experiments.}
    
    \label{tab:datasplits}
    \setlength{\tabcolsep}{5.5pt}
    \begin{NiceTabular}{@{}ccc}%[
        \toprule
        \header{Eval} & \header{Train/test} & \header{Identifier} \\\midrule
        \xmark & \cmark & run\_2024\_05\_09\_16\_18\_00 \\
        \xmark & \cmark & run\_2024\_05\_09\_16\_23\_24\\
        \xmark & \cmark & run\_2024\_25\_09\_16\_31\_52 \\
        \xmark & \cmark & run\_2024\_25\_09\_16\_35\_01 \\
        \xmark & \cmark & run\_2024\_25\_09\_13\_16\_38 \\
        \xmark & \cmark & run\_2024\_25\_09\_16\_15\_52 \\
        \xmark & \cmark & run\_2024\_25\_09\_16\_20\_50 \\
        \xmark & \cmark & run\_2024\_26\_09\_15\_50\_44 \\
        \xmark & \cmark & run\_2024\_15\_10\_11\_37\_02 \\
        \xmark & \cmark & run\_2024\_15\_10\_15\_21\_49 \\
        \xmark & \cmark & run\_2024\_15\_10\_15\_23\_39 \\
        \xmark & \cmark & run\_2024\_15\_10\_11\_32\_12 \\
        \xmark & \cmark & run\_2024\_15\_10\_11\_30\_20 \\\midrule
        \cmark &  \xmark  & run\_2024\_17\_10\_14\_21\_45 \\
        \cmark &  \xmark  & run\_2024\_17\_10\_14\_25\_17 \\
        \cmark &  \xmark  & run\_2024\_17\_10\_14\_29\_11 \\
        \cmark &  \xmark  & run\_2024\_17\_10\_14\_31\_58 \\
        \cmark & \xmark  & run\_2024\_17\_10\_14\_36\_44 \\
        \cmark & \xmark  & run\_2024\_17\_10\_14\_39\_39 \\
    \bottomrule
    \end{NiceTabular}
\end{table}

\begin{figure}[h!]
    \centering
    \includegraphics[width=\columnwidth]{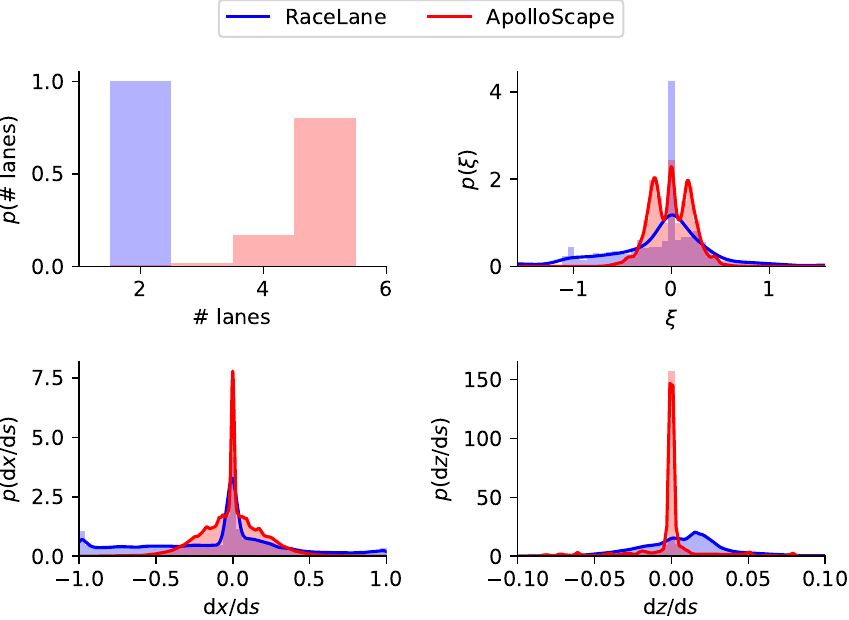}
    
    \caption{Empirical distributions of labels in \trilanes (camera 0, training data) vs. Apollo (standard split, training data).}
    \label{fig:datadistributions}
\end{figure}

\newpage\textbf{Statistics.} To illustrate the data distribution of the \trilanes labels, we provide a set of plots that compare statistics such as the distribution of: the number of lanes in each camera image; the normalized turn of the lane
\begin{equation}
\xi = \text{tan}^{-1}\Big(\frac{y(-1) - y(0)}{x(-1)-x(0)}\Big)
\end{equation}
where $(x(0),y(0))$ denotes the first point of the lane label, and $(x(-1),y(-1))$ denotes the last point on the lane label in a 2D BEV projection in $\{\mathcal{R}\}$; and the change of the $y(s), z(s)$ coordinates of the lane label as a function of a path-distance $s$ (m). The distributions are shown in~\cref{fig:datadistributions} for \trilanes and ApolloScape. This highlights differences in \trilanes compared to regular AD driving datasets such as ApolloScape, motivating its creation for assessing 3D lane detection in racing applications. In particular,  $p(\der y/\der s)$ and  $p(\der z/\der s)$ have significantly greater support. For example, several segments of the lane are almost parallel to the $y$ direction in $\{\mathcal{R}\}$, i.e., $\der y/\der s\equiv \pm 1$, with the tangent of the lane boundary being almost orthogonal to the vehicle's direction of movement. This is not unusual during tight turns in racing, but it is uncommon in the labels of the regular driving datasets~\cite{huang2020apollo}.

\begin{figure}
    \centering

    \includegraphics[width=\columnwidth]{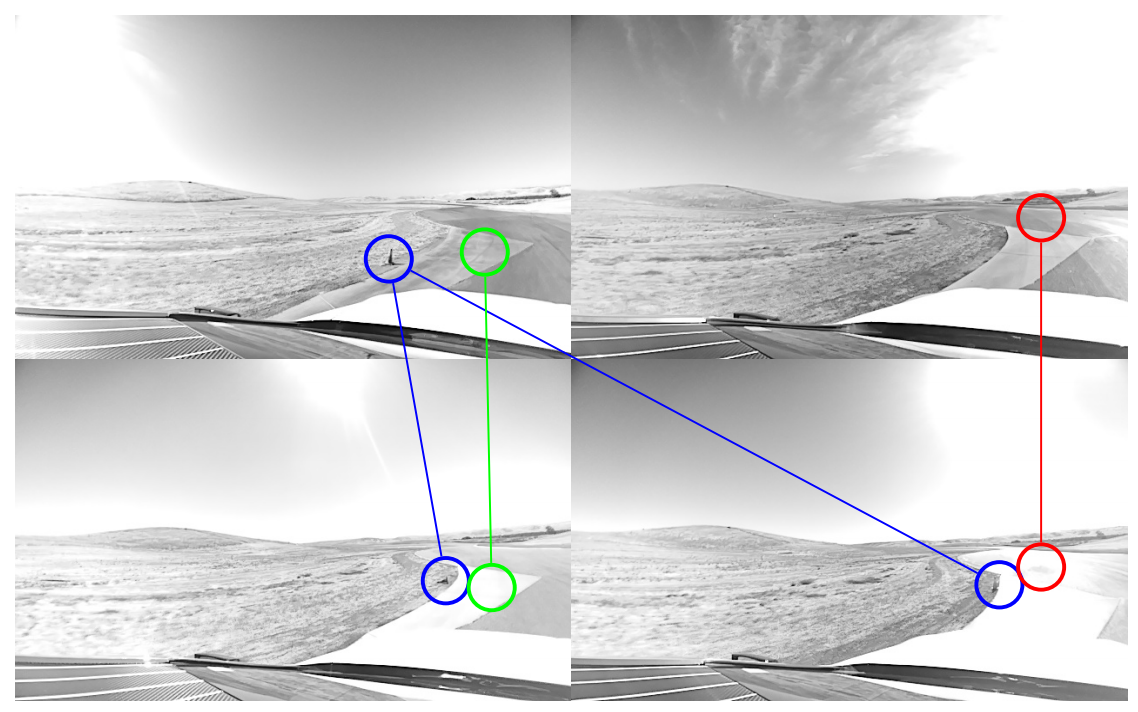}    
    \caption{Variability between the runs in \trilanes: brightness (green), sun glare (red), and apex cone location/existence (blue).}
    \label{fig:datavariatoins}
\end{figure}

\textbf{Variations across runs.} There is significant variation in the images of \trilanes cross runs, as shown in.~\cref{fig:datavariatoins}. Notably, the brightness of the road changes throughout the day (green), there is sun glare on the asphalt and car roof (red), the distribution of grass and dirt on the roadside changes in time, and the apex cones are sometimes present, sometimes knocked over, and missing entirely in other runs (blue).

\section{Training Details}\label{app:training}

\begin{figure}[h!]
    \centering
    \includegraphics[width=\columnwidth]{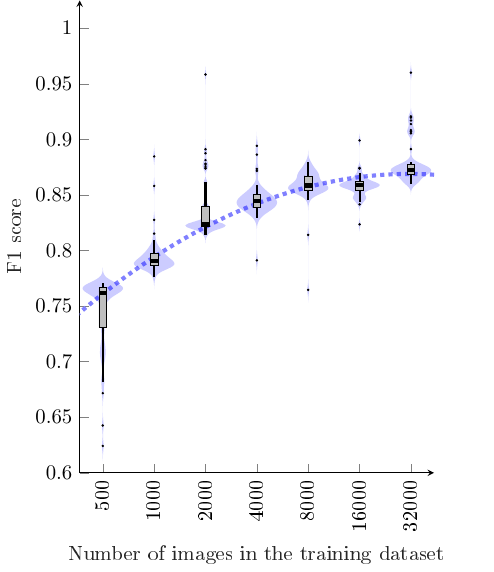}
    
    \caption{F1 classification score when training camera 1 on increasingly large subsets of \trilanes. The score is computed for the best model over 50 training epochs. A quadratic fit to the median shows that performance plateaus after 20-30k images. This experiment used a BEV perspective with $x\in[3,103]$, and the absolute scores are slightly lower than those reported in in~\cref{sec:results}.}
    \label{fig:compscaling}
\end{figure}

We use \bev as originally defined in~\cite{wang2023bev}, but with the BEV perspective in $\{\mathcal{R}\}$ as a plane with $x\in [0, 100]$ and 
$y\in [-12, 12]$. We use a grid size of $0.5$m such that each slice of the BEV output tensor consists of 9600 pixels. The virtual camera size is the same as the input image size, with $640\times 400$ pixels. Each network is trained over 50 epochs with the AdamW optimizer with a learning rate of $10^{-3}$ and weight decay of $10^{-2}$. The best model for the evaluations is chosen based on the F1 classification score. We nominally use a random subset of $2\cdot 10^4$ images per camera from the train split. This is due to a significant improvement when training on 500 to 20k images with a single camera, and diminishing returns after 20k images (see Fig.~\ref{fig:compscaling}).

\begin{figure}[h!]
\centering
    \includegraphics[width=0.88\columnwidth]{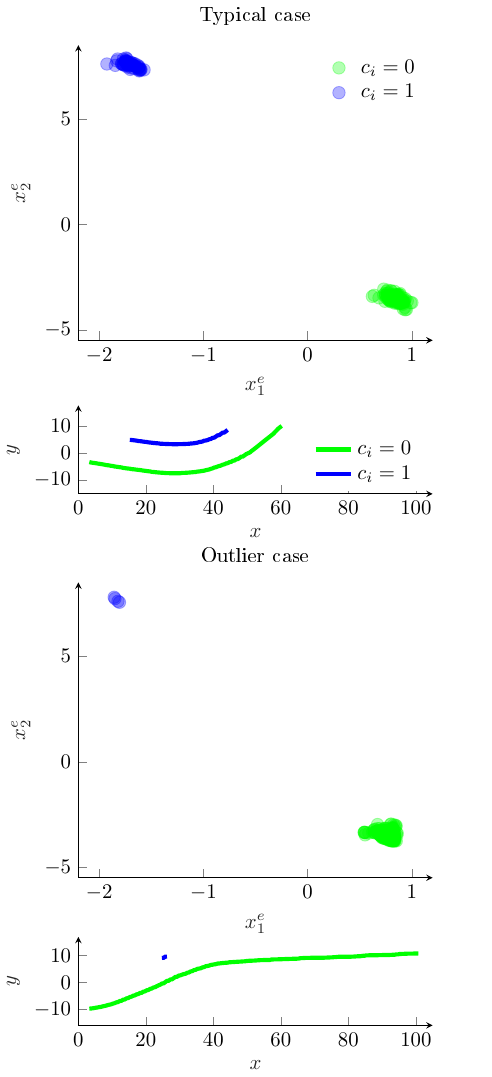}
    
    \caption{Predictions with the modified \bev. \textit{Left:} Typical bimodal cluster with $\sum_ic_i \approx \sum_i(1-c_i)$.  \textit{Right:} Long-tail event with a bimodal cluster where$\sum_ic_i \ll \sum_i(1-c_i)$, handled by line 9 in \cref{alg:clustering}. The top subplots correspond to the bottom subplots, shown in a BEV projection in the road frame.}
    \label{fig:clusteringqualitative}
\end{figure}

\section{Clustering Method}\label{app:clustering}
In the context of racing, we have a strong prior that there will never be more than two lane boundaries visible in any prediction (see~\cref{fig:datadistributions}). Therefore, it is sufficient to partition the candidate feature vectors into (at most) two disjoint sets. The method proposed in~\cref{sec:method:adaptations} does this efficiently and handles the cases of two separable clusters and a single cluster. However, special cases occur in $\lesssim 0.1\%$ of predictions, where one cluster contains drastically fewer points than the other cluster. This special case can be handled by checking the absolute distance of the resulting clusters along the principal component, which is computed automatically when checking the conditional statement on $d_i\triangleq \bar\vvec^{\top}(\xvec_i^e-\mvec)$ for all $\xvec^e\in\mathcal{X}$ in~\eqref{eq:clusterlogic}. By comparing the distance $\delta_k = \max(\{|d_i-d_j|\;;\;c_i=c_j=k\}$ to the principal component $\bar\lambda$, we can assess if the clustering corresponds to this long-tail event, and re-cluster based on the half-plane $0=\bar\vvec^{\top}(\xvec_i^e - (\mvec + \tfrac12(\delta_0 + \delta_1)\bar\vvec))$ instead. If there are $n$ points to be clustered, the resulting algorithm is $O(D_{\mathrm{out}}^3)$ where $m$ is the dimension of the features $\xvec_i^e$. As $D_{\mathrm{out}}=2$, the dominating term is the computation of the covariance matrix, which is linear in the number of points $O(|\mathcal{X}|)$.

Two sample predictions are shown in ~\cref{fig:clusteringqualitative} with two sample predictions, and a sketch of the clustering algorithm is provided in~\cref{alg:clustering}and~\cref{alg:cluster}. On 500 randomly sampled predictions on the \trilanes dataset, the computational time of clustering as in~\cite{wang2023bev} is 7.8ms ($1\sigma$ 2.5) whereas the computational time with \cref{alg:clustering} is 0.65ms (1$\sigma$ 0.27), one order of magnitude less. In practice, we observe a 10-20x speed up (see Tab~\ref{tab:optimization}).

\begin{algorithm}[h]
\caption{Fast lane clustering for uni- or bimodal clusters.}
\label{alg:clustering}
\begin{algorithmic}[1]
\State \textbf{receive} $\mathcal{X}$ and $\zeta$
\State Compute $\mvec$ and $\C$
\State Compute $(\bar\lambda, \bar\vvec,\ubar\lambda)$ from $\C$
\If{$\bar\lambda/\ubar{\lambda} < \zeta$}
\State $\cvec = \Z$
\Else
\AlgCommentIndent{Bimodal clustering}
\State $\cvec, \dvec = \texttt{cluster}(\mathcal{X}, \bar\vvec, \bar\vvec^{\top}\mvec)$
\State $\delta_k = \max(\dvec(\cvec = k))-\min(\dvec(\cvec = k))$
\AlgComment{Handle long-tail event}
\If{$\max_k\delta_k > \sqrt{\bar\lambda}$}
\State $\cvec, \dvec = \texttt{cluster}(\mathcal{X}, \bar\vvec, \bar\vvec^{\top}\mvec+(\delta_0 + \delta_1)/2)$
\EndIf
\EndIf
\State \textbf{return} $\cvec$
\end{algorithmic}
\end{algorithm}

\begin{algorithm}[h]
\caption{The \texttt{cluster} subroutine.}
\label{alg:cluster}
\begin{algorithmic}[1]
\State \textbf{receive} $\mathcal{X}, \avec, b$
\For{$\xvec_i\in\mathcal{X}$}
\State $[\dvec]_i = \avec^{\top}\xvec_i^e - b$
\State $[\cvec]_i = ([\dvec]_i>0)\;:\; 1 \;? \;0$
\EndFor
\State \textbf{return} $\cvec,\dvec$
\end{algorithmic}
\end{algorithm}

\section{Regression}
In this section, we give additional details on how the regression problem is formulated and solved. We discuss how the regularizer is implemented and the closed-form solution to the unconstrained minimization of the loss $J(\pvec)$ ~\eqref{eq:regressionloss} in~\cref{app:curvature}. We then describe how the transform and cluster buffers are used in the multi-camera regression in~\cref{app:SE3}. 

\subsection{B\'ezier Curves, Total Variation, and Moments}\label{app:curvature}
The squared $\ell_2$-norm of a polynomial is convex in its coefficients~\cite{de1978practical}. To see this, and without loss of generality, consider two one-dimensional B\'ezier curves with control points $\pvec = (p_0;\cdots;p_m)\in\Real^{m+1}$ and $\qvec = (q_0;\cdots;q_n)\in\Real^{n+1}$, with degrees $m$ and $n$ respectively. Then
\begin{equation}
I=\int \B(\lambda;\pvec) \cdot \B(\lambda;\qvec) \der \lambda = \sum_{i=0}^m\sum_{j=0}^n p_iq_j\begin{pmatrix}m\\i
\end{pmatrix}\begin{pmatrix}n\\j
\end{pmatrix}I_{ij}^{mn},
\end{equation}
where $I_{ij}^{mn} = \int\alpha_i^m(\lambda)\alpha_j^n(\lambda)\der \lambda$ and the dependency in $\lambda$ is isolated in $\alpha_i^n(\lambda) = (1-\lambda)^{(n-i)}\lambda^i$. We find the integral as
\begin{equation}
I_{ij}^{mn} = \frac{\Gamma(i+j+1)\Gamma(m+n-i-j+1)}{\Gamma(m+n+2)}.
\end{equation}
and therefore $I = \pvec^{\top}\Q_{mn}\qvec$ with elements
\begin{equation}
    [\Q_{mn}]_{i+1,j+1} = \frac{m!n!(i+j)!(m+n-i-j)!}{i!j!(m-i)!(n-j)!(m+n+1)!}.
\end{equation}
As such, for the curve $\B(\lambda;\pvec)$ of degree $n$ we can write the squared $\ell_2$-norm of the curvature of a lane as a quadratic 
\begin{equation}\label{eq:app:weakconvexity}
\int\|\B(\lambda; \pvec)\|_2^2\der\lambda = \pvec^{\top}\Q_{nn}\pvec,
\end{equation}
and convexity follows as integrating an integrable positive function yields a positive integral, thus $\pvec^{\top}\Q_{nn}\pvec \geq 0$. 

\textbf{The TV Regularizer}
The derivative of a Bezi\'er curve of degree $n$ is a Bezi\'er curve of degree $n-1$. Specifically, if $(\der^k/\der\lambda^k)\B(\lambda; \pvec) = \B(\lambda; \qvec)$, then $\qvec = \M_{k}^{n} \pvec$, where
\begin{equation}
\M_{1}^n = \begin{bmatrix}
1 & -1 & 0 & \cdots & 0 & 0\\
0 & 1 & -1 & \cdots & 0 & 0\\
\vdots &  & & \ddots &  & \vdots\\
0 & 0 & 0 &  & -1 & 0 \\
0 & 0 & 0 & \cdots & 1 & -1
\end{bmatrix}\in\Real^{n-1\times n},
\end{equation}
and $\M_{k}^{n} = \M_{1}^{n-k}\cdots\M_{1}^{n-1}\M_{1}^{n}$. Consequently, we can express the proposed TV regularizer for a one-dimensional Bezi\'er curve as a quadratic function in $\pvec$, as
\begin{subequations}
\begin{align}
    \sum_{k=1}^n \hspace{-1pt}\beta_k&\hspace{-3pt}\int\hspace{-3pt}\|\B^{(k)}(\lambda; \pvec)\|_2^2\der\lambda =\\
    &=\pvec^{\top}\underbrace{ \Big(\sum_{k=1}^n\beta_k(\M_{k}^n)^{\top}\Q_{n-k,n-k}\M_{k}^n\Big) }_{\triangleq \Q^{\mathrm{1D}}(\betabf)}\pvec.
\end{align}
\end{subequations}
where $\Q^{\mathrm{1D}}(\betabf)\succeq 0$ by the weak convexity of~\eqref{eq:app:weakconvexity}, and $\betabf = (\beta_1,...,\beta_n)$ is a vector of the regularizer weights.

\textbf{The Estimator.} To find the minimizer of $J(\pvec)$ in~\eqref{eq:regressionloss}, we note that for a $d=3$-dimensional Bezi\'er curve, the TV-term in $J$ can be written as a quadratic $\pvec^{\top}\Q(\betabf)\pvec$ with $\Q(\betabf)=\I_3\otimes \Q^{\mathrm{1D}}(\betabf)$. As $J$ is quadratic, we have that
\begin{equation}\label{app:minimizer}
\hat\pvec \hspace{-1pt}=\hspace{-1pt} \Big(
    \Q(\betabf) + \sum_{i=1}^m\A_i^{\top}\Sigmabf_i^{-1}\A_i
\Big)^{-1}
 \Big(
\sum_{i=1}^m\A_i^{\top}\Sigmabf_i^{-1}\yvec_i
 \Big)\hspace{-1pt},\hspace{-1pt}
\end{equation}
where we let $\B(\lambda_i; \pvec) = \A_i \pvec$ using the definition in~\eqref{eq:bezdef}.

\textbf{Bias.} For simplicity, we let $\tilde\pvec = \pvec - \hat\pvec$ be the estimation error, and write the estimator in~\eqref{app:minimizer} in the matrices
\begin{subequations}
\begin{align}
\bar\Sigmabf &= \text{blkdiag}(\Sigmabf_1, ..., \Sigmabf_m)\\
\bar\A &= (\A_1, ... ,\A_m)\\
\bar\yvec &= (\yvec_1; ... ;\yvec_m)\\
\bar\epsbf &= (\epsbf_1; ... ;\epsbf_m)
\end{align}
\end{subequations}
such that $\bar\yvec = \bar\A\pvec + \bar\epsbf$, where then
\begin{equation}
\hat\pvec=(\Q(\betabf) + \bar\A^{\top}\bar\Sigmabf^{-1}\bar\A)^{-1}\bar\A^{\top}\bar\Sigmabf^{-1}\bar\yvec.
\end{equation}
\begin{subequations}
As $\M = \Q(\betabf) + \bar\A^{\top}\bar\Sigmabf^{-1}\bar\A$ is invertible, the bias is
\begin{align}
\mathbb{E}[\tilde\pvec] &= \pvec -\M^{-1}\bar\A^{\top}\bar\Sigmabf^{-1}\bar\A \pvec\Rightarrow\\
\M\mathbb{E}[\tilde\pvec] &= \M\pvec - \bar\A^{\top}\bar\Sigmabf^{-1}\bar\A\pvec=\Q(\betabf)\pvec\Rightarrow\\
\mathbb{E}[\tilde\pvec] & =\M^{-1}\Q(\betabf)\pvec
\end{align}
\end{subequations}
Thus, the regularizer introduces a weak bias that increases with the tuning $\beta_i$.

\subsection{Synthesizing the Transforms}\label{app:SE3}
The exponential map of $\mathsf{SE(3)}$ is implemented using Rodrigues' formula~\cite{murray2017mathematical}. With $\xibf = (\omegabf; \vvec)\in\Real^6$ and
\begin{equation*}
[\xibf]^{\land}_{\mathsf{SE(3)}} = 
\begin{bmatrix}
[\omegabf]^{\land}_{\mathsf{SO(3)}} & \vvec \\ \Z & 0
\end{bmatrix}, \quad 
[\xibf]^{\land}_{\mathsf{SO(3)}} = 
\begin{bmatrix}
 0 & \omega_3  & \omega_2 \\ 
\omega_3 & 0 & \omega_1 \\  
\omega_2 & \omega_1 & 0 
\end{bmatrix},
\end{equation*}
spanning the relevant Lie algebras, we have that
\begin{subequations}
\begin{align}
\T(\omegabf, \vvec) &= \exp([\xibf]^{\land}_{\mathsf{SE(3)}}) = 
\begin{bmatrix}
\R(\omegabf) & \A(\omegabf)\vvec \\ \Z & 1
\end{bmatrix},\\
\R(\omegabf)&=\I +
f(\eta)[\omegabf]_{\mathsf{SO(3)}}^{\land} + g(\eta)\
([\omegabf]_{\mathsf{SO(3)}}^{\land})^2,\\
\A(\omegabf) &=\I +g(\eta) [\omegabf]_{\mathsf{SO(3)}}^{\land}
+ h(\eta)([\omegabf]_{\mathsf{SO(3)}}^{\land})^2,
\end{align}
\end{subequations}
where $\eta = \|\omegabf\|_2\notag$, and the functions
\begin{subequations}\label{eq:taylorexpmap}
\begin{align}
\hspace{-3pt}f(\eta) &= \frac{\sin(\eta)}{\eta}  \hspace{26pt}= 1   - \frac{\eta^2}{6}   + \frac{\eta^4}{120} + o(|\eta|^6),\\
\hspace{-3pt}g(\eta) &= \frac{(1-\cos(\eta))}{\eta^2} = \frac12 - \frac{\eta^2}{24}  + \frac{\eta^4}{720} + o(|\eta|^6),\\
\hspace{-3pt}h(\eta) &= \frac{(\eta-\sin(\eta))}{\eta^3} = 
\frac16 - \frac{\eta^2}{120} + \frac{\eta^4}{560} + o(|\eta|^6),
\end{align}
\end{subequations}
are evaluated using exactly when $|\eta| > 10^{-15}$ and by the Taylor expansions about $\eta=0$ in~\eqref{eq:taylorexpmap} when $|\eta|$ is small. Now, assume that we measure a sequence $\{(t_i,\omegabf_i,\vvec_i)\in\Real^7:t_{i+1}>t_i\}_{i=k}^K$. To compute the transform relating the road frames $\{\mathcal{R}\}$ at $t_k$ and $t_K$ we chain the transforms
\begin{equation}\label{eq:clustertransforms}
\T_{K|k} = \prod_{i=k}^{K-1} \T(\Delta_i\omegabf_i, \Delta_i\vvec_i)\in\mathsf{SE(3)},
\end{equation}
where  $\Delta_i = t_{i+1}-t_i\geq 0$. We store the intermediary transforms $\T(\Delta_i\omegabf_i, \Delta_i\vvec_i)\in\mathsf{SE(3)}$ in a circular buffer, and compute these each time an IMU or odometry measurement is taken, in order to distribute computations over time. Similarly, each time a new cluster prediction is received, we compute transforms~\eqref{eq:clustertransforms} relating consecutive cluster predictions. For the regression, we only need to go through the transforms relating the various clusters in the second buffer in order to project the clusters into the same road frame.

\newcommand\ellbf{\boldsymbol{\ell}}
\section{Metrics}\label{app:metrics}
In the 3D lane detection literature, there are various metrics used for evaluating performance. To clarify these metrics, we here define a lane $\ellbf:[0,1]\mapsto \Real^3$. We let $\hat\ellbf_j$ be the $j$th lane estimate and $\ellbf_i$ denote the $i$th ground truth lane. In practice, the ground truth lanes $\ellbf_i$ are simply a collection of points associated with $\Lcal_i = \{\lambda_k\in[0,1]\}$. As such, evaluating the estimated lane $\hat\ellbf_j(\lambda_k)$ for any $\lambda_k\in\Lcal_i$ amounts to an interpolation. Thus, we need to associate each $\{\ellbf_i\}_i$ with $\{\hat\ellbf_j\}_j$. To this end, we follow~\cite{guo2020gen} and let
\begin{align*}
c_{ij} &= \min(\|\ellbf_i(\lambda) - \hat\ellbf_j(\lambda)\|_2, \epsilon),\\
d_{ij}&=\sum_{\lambda\in\Lcal_i}
\begin{cases}
c_{ij} & \mathrm{if}\;\;\Ical_{\ellbf_i}(\lambda)=\Ical_{\hat\ellbf_j}(\lambda)=1\hspace{10pt}\text{{\color{gray}// Both vis.}}\\
0 & \mathrm{if}\;\;\Ical_{\ellbf_i}(\lambda)\land \Ical_{\hat\ellbf_j}(\lambda)=0 \hspace{10pt}{\color{gray}\text{// Neither vis.}}\\
\epsilon &  \mathrm{otherwise}\hspace{63pt}{\color{gray}\text{// One vis.}}\\
\end{cases},
\end{align*}
where $d_{ij}$ is a measure of how close two lanes are, and $\Ical_{\ellbf}\;:\;[0,1]\mapsto \{0,1\}$ is a map indicating if a given point on the lane $\ellbf$ is visible at a given $\lambda$. We can then formulate the bipartite matching problem as a min-cost-flow problem, and associate each lane $\ellbf_i$ with a particular lane $\hat\ellbf_j$ while minimizing the sum of associated $d_{ij}$ costs. The min-cost-flow solution returns lane assignments as tuples $(i,j)$. For each such tuple, we compute
\begin{align}
\mathrm{P}_{ij} &= \frac{\sum_{\lambda\in\Lcal_i} (\|\ellbf_i(\lambda) - \hat\ellbf_j(\lambda)\|_2 < \epsilon)}{\sum_{\lambda\in\Lcal_i}I_{\ellbf_i}(\lambda)},\\%{\color{gray}\hspace{20pt}\text{//}\Bigg(\frac{\text{All interpolated EST that are close to GT}}{\text{All visible points in GT (TP+FN)}}\Bigg)}\\
\mathrm{R}_{ij} &= \frac{\sum_{\lambda\in\Lcal_i} (\|\ellbf_i(\lambda) - \hat\ellbf_j(\lambda)\|_2 < \epsilon)}{\sum_{\lambda\in\Lcal_i}I_{\hat\ellbf_j}(\lambda)}.%{\color{gray}\hspace{20pt}\text{//}\Bigg(\frac{\text{All interpolated EST that are close to GT}}{\text{All visible points in EST (TP+FP)}}\Bigg)}
\end{align}
The average precision and recall measures are combined into an average F1 score. In~\cite{guo2020gen,chen2022persformer,garnett20193d,pittner2024lanecpp,luo2023latr}, these values are truncated, in the sense that $\Pcal(\alpha)=\{\mathrm{P}_{ij}\in[0,1]|\mathrm{P}_{ij}>\alpha\}$ and $\mathcal{R}(\alpha)=\{\mathrm{R}_{ij}\in[0,1]|\mathrm{R}_{ij}>\alpha\}$, where
\begin{align}
\text{AP} &= \frac{1}{|\Pcal(\alpha)|}\sum_{p\in \Pcal(\alpha)}p,\\
\text{AR} &= \frac{1}{|\mathcal{R}(\alpha)|}\sum_{r\in \mathcal{R}(\alpha)}r,\\
\text{F1} &= 2\frac{\mathrm{AP}\cdot\mathrm{AR}}{\mathrm{AP}+\mathrm{AR}}.\label{eq:averaging}
\end{align}
This is the definition of the F1 score reported in~\cite{guo2020gen,chen2022persformer,garnett20193d,pittner2024lanecpp,luo2023latr}, and the metric is notably a function of the tolerances $\epsilon$ and $\alpha$. From the benchmark evaluation codes of~\cite{guo2020gen} which subsequent cited methods use, these are set to $\epsilon=1.5$m, and $\alpha=0.75$, and we therefore use the same tolerances. The metrics are reported as \textbf{F1}, \textbf{AP}, and \textbf{AR}, respectively.

We also report mean absolute errors (MAEs) in the lane predictions over the visible portions of matched lanes. That is, if we have a ground truth and estimate lane tuple $(i,j)$ from the assignment problem, and we consider a set of points in the near field $[0,x^{-}]$ and in the far field $[0,x^{+}]$, then we can let $\Lcal_i^{\pm}=\{\lambda\in\Lcal_i\::\: [\ellbf_i(\lambda)]_1\in[0,x^{\pm}]\}$ and compute
\begin{align}\label{eq:abserr}
    \text{abserr}^{\pm}_{k}=|\Lcal_i^{\pm}|^{-1}\hspace{-20pt}\sum_{
    \substack{\lambda \in\Lcal_i^{\pm}\\
    \Ical_{\ellbf_i}(\lambda)=\Ical_{\ellbf_j}(\lambda)=1
    }}\hspace{-20pt}|[\ellbf_i(\lambda) - \hat\ellbf_j(\lambda)]_k|,
\end{align}
for different ranges $x^{\pm}$. The ``near'' errors are characterized by $x^{-}=40$m, and the ``far'' errors are characterized by $x^{+}=100$m, and these values are used in~\cite{guo2020gen,chen2022persformer,garnett20193d,pittner2024lanecpp,luo2023latr}. The $y$ ($k=2$) and $z$ ($k=3$) errors are presented separately and computed over the entire test dataset as $\text{MAE}_k^{\pm} = \mathbb{E}[\text{abserr}^{\pm}_{k}]$. Contrary to the previous works, we report the MAEs as \textbf{Y (near)}, \textbf{Y (far)}, \textbf{Z (near)} and \textbf{Z (far)}, respectively, as this better aligns with our road frame definition.

\section{Experiments}\label{app:datasplits}

For the training/testing, we use an 80/20 split on a subset of the runs in \trilanes. As the cameras are sampled at high rates, for any image in the test set, there is likely a similar image in the train set. The metrics of interest are therefore computed on a different set of runs (see Tab~\ref{tab:datasplits}). These evaluation runs are further outside the train/test data distribution, and metrics on these runs are therefore more indicative of how the system is likely to perform in deployments.

\subsection{Monochromatic images}\label{app:mono}
As \trilanes consists of grayscale images, we assess the impact of using monochromatic images in various detectors in PersFormer\cite{chen2022persformer},~\cite{luo2023latr},~\cite{wang2023bev}. For this purpose, we use ApolloScape~\cite{huang2020apollo} and simply average the RGB channels. The test results on the ``standard'' split are shown in~\cref{tab:monocheck}, indicating a marginal performance decrease across all methods and superior classification results with \bev. While it is known that \bev yields similar performance to LATR, it is surprising that moving to gray-scale images has such a marginal impact on performance. This is the reason that we use monochromatic images in \trilanes, to conserve bandwidth when sampling images at high rates.

\begin{table}[t!]
    \centering
    %\resultQ{Table Showing performance with Persformer, LATR, Bev-Lane-Det, ideally on Apollo Mono Data.}
    \caption{\textbf{Gray-scale ablation, Apollo Dataset:} GS--apply a gray-scale transform to inputs. Metrics are the same as in~\cite{guo2020gen}.}
    
    \label{tab:monocheck}
        \setlength{\tabcolsep}{5.5pt}
        \resizebox{\columnwidth}{!}{\begin{NiceTabular}{@{}ccccccccc}[
            code-before =%
            % Best
            \rectanglecolor{\bestcolor}{6-3}{6-5}
            \rectanglecolor{\bestcolor}{7-6}{7-6}
            \rectanglecolor{\bestcolor}{7-7}{7-7}
            \rectanglecolor{\bestcolor}{5-8}{5-8}
            \rectanglecolor{\bestcolor}{5-9}{5-9}
            % Second
            \rectanglecolor{\secondbestcolor}{7-3}{7-5}
            \rectanglecolor{\secondbestcolor}{5-6}{5-6}
            \rectanglecolor{\secondbestcolor}{5-7}{5-7}
            \rectanglecolor{\secondbestcolor}{4-8}{4-8}
            \rectanglecolor{\secondbestcolor}{2-8}{2-8}
            \rectanglecolor{\secondbestcolor}{6-8}{6-8}
            \rectanglecolor{\secondbestcolor}{4-9}{4-9}
        ]
        \toprule
        {\header{Method}} & {\header{GS}} & {\header{F1} $\uparrow$} & {\header{AP} $\uparrow$} & {\header{AR} $\uparrow$} & {\header{Y (near)}} & {\header{Y (far)}} & {\header{Z (near)}} & {\header{Z (far)}} \\ \midrule
        \header{PersFormer}~\cite{chen2022persformer} & {\large\xmark} & 91.87 & 92.00 & 91.75 & 0.059 & 0.3760 & 0.011 & 0.232 \\
        \header{PersFormer}~\cite{chen2022persformer} & {\large\cmark} & 90.82 & 90.47 & 91.23 & 0.060 & 0.384 & 0.012 & 0.239 \\
        \header{LATR}~\cite{luo2023latr} & {\large\xmark} & 96.32 & 0.955 & 0.973 & 0.034 &  0.284 & 0.011 & 0.212 \\
        \header{LATR}~\cite{luo2023latr} & {\large\cmark} & 96.76 & 0.974 &0.959 & 0.029 &0.266 &0.008 &0.203 \\
        \header{BevLaneDet}~\cite{wang2023bev} & {\large\xmark} & 97.70 & 0.977 & 0.977 & 0.059 & 0.376 & 0.011 & 0.234 \\
        \header{BevLaneDet}~\cite{wang2023bev} & {\large\cmark} & 97.50 & 0.974 & 0.976 & 0.028 & 0.251 & 0.027 & 0.229  \\
        \bottomrule
    \end{NiceTabular}}
\end{table}

\subsection{Multi-Camera Evaluation} \label{app:ensemble:training}
In \cref{tab:multicamerapredictions} each ``camera set model'' is trained on data from cameras in the given set. To make this comparison fair, each model is trained on the same total number of images. For example, the models associated with camera 0 and camera 1 are trained on $N$ unique images each; the model associated with the camera set $\{0,1\}$ is trained on a random subset of $N/2$ images from each camera. Similarly, the model associated with $\{0,1,2,3\}$ is trained on $N/4$ images randomly sampled from the training data associated with each camera.
 
\subsection{Model Optimization}\label{app:modeloptimization}
The model optimization ablations were done with a model trained on the train split of \trilanes, including data from September and October, up to and including October 15th (see 1st split in~\cref{tab:datasplits}). The computation times for the 3070 were taken while deploying this model in the vehicle, and the compute times on the 4070 were also computed in the vehicle, but when stationary in the garage. The clustering computation times were averaged over the 3070 and 4070 runs, but did not change meaningfully as the clustering is run on the CPU. The metrics were computed based on single-camera predictions with the 4070 GPU, with respect to the ground truth lanes visible in each camera individually. In the model optimizations with HP and PTQ, the models were compiled separately for each GPU to maximize performance. A complete summary of the metrics is provided in~\cref{tab:app:regressionoptimization}, where the average metrics are computed over all the cameras (one model per camera).

\begin{table}[t!]
    \centering
    \caption{\textbf{Optimization Ablations (multi-camera regressed output):} HP -- Half Precision, PTQ -- Post Training Quantization, PCA -- Use the PCA clustering.  Metrics computed on the evaluation dataset.}
    
    \label{tab:app:regressionoptimization}
    \setlength{\tabcolsep}{5.5pt}
    \resizebox{\columnwidth}{!}{\begin{NiceTabular}{@{}c|ccc|ccccccc}[
    code-before =%
            \rectanglecolor{gray!20}{4-1}{4-11}%
            \rectanglecolor{gray!20}{6-1}{6-11}%
            \rectanglecolor{gray!20}{8-1}{8-11}%
            \rectanglecolor{gray!20}{10-1}{10-11}%
            \rectanglecolor{gray!20}{12-1}{12-11}%
            \rectanglecolor{gray!20}{14-1}{14-11}%
            \rectanglecolor{gray!20}{16-1}{16-11}%
            \rectanglecolor{gray!20}{18-1}{18-11}%
            \rectanglecolor{gray!20}{20-1}{20-11}%
            \rectanglecolor{gray!20}{22-1}{22-11}%
            \rectanglecolor{gray!20}{24-1}{24-11}%
            \rectanglecolor{gray!20}{26-1}{26-11}%
            \rectanglecolor{gray!20}{28-1}{28-11}%
            \rectanglecolor{gray!20}{30-1}{30-11}%
            \rectanglecolor{gray!20}{32-1}{32-11}
    ]
        \toprule
        \header{Cameras} & \multicolumn{3}{c}{\header{Optimization}} & \multicolumn{7}{c}{\header{Metrics}}\\
        & \header{HP}  &
        \header{PTQ} & 
        \header{PCA} & \header{F1} $(\uparrow)$ & \header{AP} $(\uparrow)$  & \header{AR} $(\uparrow)$ & \header{Y (near)} & \header{Y (far)} & \header{Z (near)} & \header{Z (far)} \\
        \midrule
    0 & \xmark & \xmark & \xmark & 91.28 &       89.16 &    93.50 &            0.14 &          0.53 &            0.09 &          0.44 \\
    0 & \xmark & \xmark & \cmark & 91.10 &       89.02 &    93.29 &            0.14 &          0.53 &            0.09 &          0.44 \\
    0 & \cmark & \xmark & \xmark & 91.28 &       89.16 &    93.50 &            0.14 &          0.53 &            0.09 &          0.44 \\
    0 & \cmark & \xmark & \cmark & 91.10 &       89.02 &    93.29 &            0.14 &          0.53 &            0.09 &          0.44 \\
    0 & \cmark & \cmark & \xmark & 91.27 &       89.15 &    93.50 &            0.14 &          0.53 &            0.09 &          0.44 \\
    0 & \cmark & \cmark & \cmark & 91.09 &       89.00 &    93.27 &            0.14 &          0.53 &            0.09 &          0.44 \\
    \midrule
    1 & \xmark & \xmark & \xmark &       92.32 &       90.00 &    94.77 &            0.13 &          0.54 &            0.08 &          0.46 \\
    1 & \xmark & \xmark & \cmark &       92.29 &       89.96 &    94.75 &            0.14 &          0.54 &            0.08 &          0.46 \\
    1 & \cmark & \xmark & \xmark &       92.32 &       90.00 &    94.77 &            0.13 &          0.54 &            0.08 &          0.46 \\
    1 & \cmark & \xmark & \cmark &       92.29 &       89.96 &    94.75 &            0.14 &          0.54 &            0.08 &          0.46 \\
    1 & \cmark & \cmark & \xmark &       92.32 &       90.00 &    94.77 &            0.13 &          0.54 &            0.08 &          0.46 \\
    1 & \cmark & \cmark & \cmark &       92.30 &       89.97 &    94.75 &            0.14 &          0.54 &            0.08 &          0.46 \\
    \midrule
    2 & \xmark & \xmark & \xmark &       95.18 &       95.45 &    94.92 &            0.15 &          0.53 &            0.10 &          0.45 \\
    2 & \xmark & \xmark & \cmark &       95.10 &       95.35 &    94.86 &            0.15 &          0.53 &            0.10 &          0.45 \\
    2 & \cmark & \xmark & \xmark &       95.18 &       95.45 &    94.92 &            0.15 &          0.53 &            0.10 &          0.45 \\
    2 & \cmark & \xmark & \cmark &       95.10 &       95.35 &    94.86 &            0.15 &          0.53 &            0.10 &          0.45 \\
    2 & \cmark & \cmark & \xmark &       95.20 &       95.46 &    94.94 &            0.15 &          0.53 &            0.10 &          0.45 \\
    2 & \cmark & \cmark & \cmark &      95.12 &       95.36 &    94.87 &            0.15 &          0.53 &            0.10 &          0.45 \\
    \midrule
    3 & \xmark & \xmark & \xmark &       95.08 &       95.31 &    94.85 &            0.15 &          0.52 &            0.10 &          0.44 \\
    3 & \xmark & \xmark & \cmark &       95.08 &       95.27 &    94.89 &            0.15 &          0.52 &            0.10 &          0.44 \\
    3 & \cmark & \xmark & \xmark &       95.08 &       95.31 &    94.85 &            0.15 &          0.52 &            0.10 &          0.44 \\
    3 & \cmark & \xmark & \cmark &       95.08 &       95.27 &    94.89 &            0.15 &          0.52 &            0.10 &          0.44 \\
    3 & \cmark & \cmark & \xmark &       95.08 &       95.31 &    94.85 &            0.15 &          0.52 &            0.10 &          0.44 \\
    3 & \cmark & \cmark & \cmark &      95.08 &       95.27 &    94.88 &            0.15 &          0.52 &            0.10 &          0.44 \\
    \midrule
    avg. & \xmark & \xmark & \xmark &       93.48 &       92.48 &    94.51 &            0.14 &          0.53 &            0.09 &          0.44 \\
    avg. & \xmark & \xmark & \cmark &       93.41 &       92.40 &    94.44 &            0.14 &          0.53 &            0.09 &          0.44 \\
    avg. & \cmark & \xmark & \xmark &       93.48 &       92.48 &    94.51 &            0.14 &          0.53 &            0.09 &          0.44 \\
    avg. & \cmark & \xmark & \cmark &       93.41 &       92.40 &    94.44 &            0.14 &          0.53 &            0.09 &          0.44 \\
    avg. & \cmark & \cmark & \xmark &       93.49 &       92.48 &    94.51 &            0.14 &          0.53 &            0.09 &          0.44 \\
    avg. & \cmark & \cmark & \cmark &      93.41 &       92.40 &    94.44 &            0.14 &          0.53 &            0.09 &          0.44 \\
    \bottomrule
    \end{NiceTabular}}
\end{table}

\begin{table}[t!]
    \centering
    \caption{\textbf{Regression Regularizer Ablations:} The weights of the squared TV regularizer in curvature, jerk, and snap are varied.}
    
    \label{tab:regression:regularizer}
    \setlength{\tabcolsep}{5.5pt}
    \resizebox{\columnwidth}{!}{\begin{NiceTabular}{@{}ccc|ccccccc}[
    code-before =%
            \rectanglecolor{gray!20}{4-1}{4-10}%
            \rectanglecolor{gray!20}{6-1}{6-10}%
            \rectanglecolor{gray!20}{8-1}{8-10}%
            \rectanglecolor{gray!20}{10-1}{10-10}%
            \rectanglecolor{gray!20}{12-1}{12-10}%
            \rectanglecolor{gray!20}{14-1}{14-10}
    ]
        \toprule
        \multicolumn{3}{c}{\header{Weights}} & \multicolumn{7}{c}{\header{Metrics}}\\
        \header{$\beta_2$}  &
        \header{$\beta_3$} & 
        \header{$\beta_4$} & \header{F1} $(\uparrow)$ & \header{AP} $(\uparrow)$  & \header{AR} $(\uparrow)$ & \header{Y (near)} & \header{Y (far)} & \header{Z (near)} & \header{Z (far)} \\
        \midrule
        0 & 0 & 0 & 88.63 &       81.86 &    96.61 &          0.19 &          0.74 &            0.20 &          0.65 \\
    $10^{-4}$ & 0 & 0 & 88.59 &       81.76 &    96.66 &            \textbf{0.18} &          0.73 &            0.20 &          0.65 \\
    $10^{-3}$ & 0 & 0 & 88.70 &       82.02 &    96.56 &            0.19 &          0.74 &            0.20 &          0.65 \\
    $10^{-2}$ & 0 & 0 & \textbf{88.77} &       82.06 &    96.66 &            0.20 &          0.74 &            0.20 &          0.65 \\
    $10^{-1}$ & 0 & 0 & 88.44 &       81.83 &    96.20 &            0.23 &          0.74 &            0.20 &          0.64 \\
    0 & $10^{-4}$ & 0 & 88.43 &       81.80 &    96.23 &            0.19 &          0.75 &            0.20 &          0.66 \\
    0 & $10^{-3}$ & 0 & 88.44 &       81.77 &    96.29 &            0.19 &          0.74 &            0.20 &          0.65 \\
    0 & $10^{-4}$ & 0 & 88.13 &       81.44 &    96.02 &            0.20 &          0.75 &            0.20 &          0.65 \\
    0 & $10^{-1}$ & 0 & 87.72 &       80.90 &    95.81 &            0.23 &          0.75 &            0.20 &          0.65 \\
    0 & 0 & $10^{-4}$ & 88.32 &       81.47 &    96.43 &            0.19 &          0.74 &            0.20 &          0.65 \\
    0 & 0 & $10^{-3}$ & \textbf{88.84} &      82.45 &    96.31 &            0.20 &          0.75 &            0.20 &          0.65 \\
    0 & 0 & $10^{-2}$ & 88.29 &       81.61 &    96.18 &            0.22 &          0.75 &            0.20 &          0.65 \\
    0 & 0 & $10^{-1}$ & 88.01 &       81.34 &    95.88 &            0.22 &          0.76 &            0.20 &          0.65 \\
    \bottomrule
    \end{NiceTabular}}
\end{table}

\subsection{Regression Ablation Studies}\label{app:regression:ablations}

To study the effect of the regularizer on the regressed 3D lane predictions, we conduct an ablation study where we sweep over the weights $(\beta_2,\beta_3,\beta_4)$. The metrics are computed for the evaluation run ``run\_2024\_17\_10\_14\_21\_45'' and reported in~\cref{tab:regression:regularizer} using the nominal \ourmethod.  These results indicate that small weight on the jerk and snap can yield marginal improvements in the classification metrics, in particular the precision. We expect this to be more impactful when the predictions are more noisy.

\newcommand\drawQualitative[5]{
\node[anchor=south west] at (4.0,{#4*2.5}) {\includegraphics[height=2.5cm]{figures/model_comparisons/#2}};
\node[anchor=south west] at (0,{#4*2.5}) {\includegraphics[height=2.5cm]{figures/model_comparisons/#1}};
\node[anchor=south west] at (6.7,{#4*2.5}) {\includegraphics[height=2.5cm]{figures/model_comparisons/#3}};
\node[rotate=90, anchor=south] at (0,{#4*2.5+1.25}) {#5};
}

\begin{figure}[t!]
\centering
\iffalse
\begin{tikzpicture}
\node[anchor=south] at (2.0,5.1) {2D image overlay};
\node[anchor=south] at (5.3,5.1) {3D view};
\node[anchor=south] at (7.4,5.1) {BEV};
\drawQualitative{overlay_persf426.png}{3d_plot_persf426.png}{xy_plot_persf426.png}{1}{PersFormer~\cite{chen2022persformer}}
\drawQualitative{overlay_bev_lane_det_og_426.png}{3d_plot_bev_lane_det_og_426.png}{xy_plot_bev_lane_det_og_426.png}{-3}{Bev-LaneDet$^{\star}$}
\drawQualitative{overlay_latr_star_426.png}{3d_plot_latr_star_426.png}{xy_plot_latr_star_426.png}{-2}{LATR$^{\star}$}
\drawQualitative{overlay_bev_lane_det_tco_1426.png}{3d_plot_bev_lane_det_tco_1426.png}{xy_plot_bev_lane_det_tco_1426.png}{-1}{Bev-LaneDet~\cite{wang2023bev}}
\drawQualitative{overlay_latr_og_426.png}{3d_plot_latr_og_426.png}{xy_plot_latr_og_426.png}{0}{LATR~\cite{luo2023latr}}
\end{tikzpicture}
\fi
\includegraphics[width=\columnwidth]{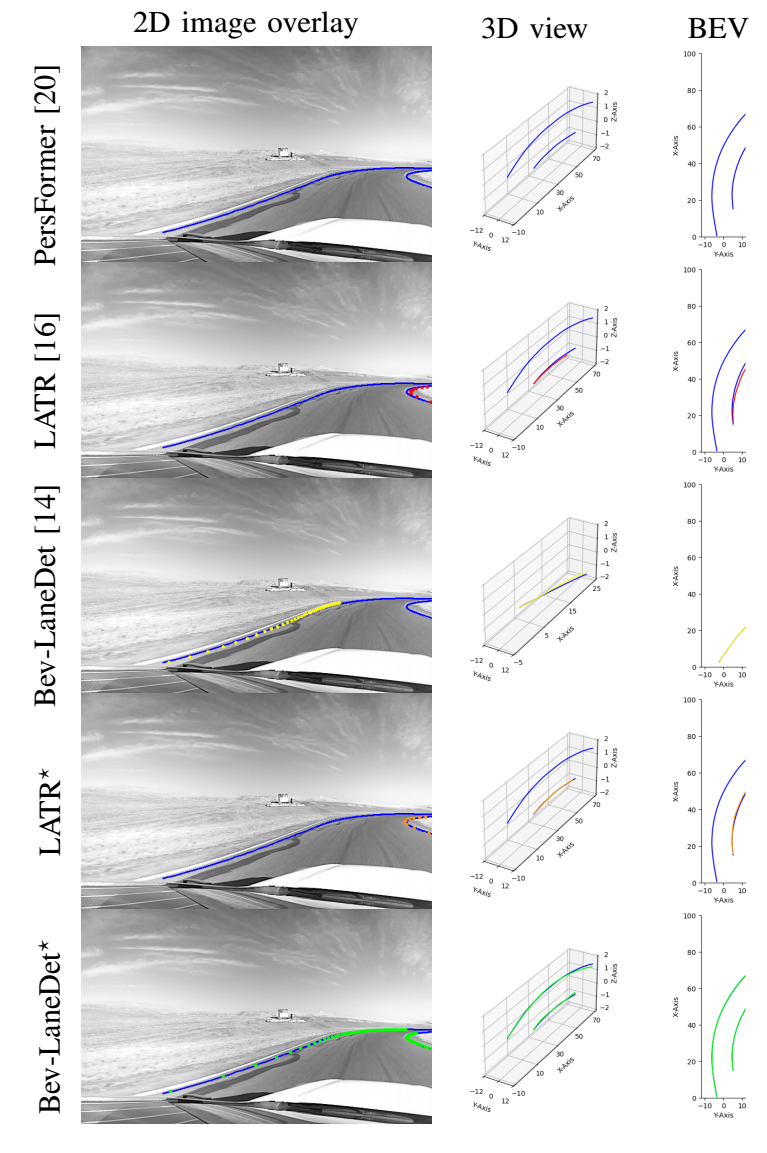}

\caption{Qualitative example of PersFormer, LATR, and Bev-LaneDet, with and without modifications, trained on a subset of \trilanes, only visualizing camera 0. PersFormer fails entirely, while original LATR (red) and Bev-LaneDet (yellow) only predict one lane. The definition of the BEV perspective differs in the Bev-LaneDet and Bev-LaneDet$^{\star}$ models, due to the lack of road-frame alignment. The modified versions of LATR (red) and Bev-LaneDet (green) predict both lanes with high accuracy (see BEV view).}
\label{fig:qualitative2D3Dfirst}
\end{figure}

\begin{figure}[t!]
\centering
\iffalse
\begin{tikzpicture}
\node[anchor=south] at (2.0,5.1) {2D image overlay};
\node[anchor=south] at (5.3,5.1) {3D view};
\node[anchor=south] at (7.4,5.1) {BEV};
\drawQualitative{overlay_persf696.png}{3d_plot_persf696.png}{xy_plot_persf696.png}{1}{PersFormer~\cite{chen2022persformer}}
\drawQualitative{overlay_bev_lane_det_og_696.png}{3d_plot_bev_lane_det_og_696.png}{xy_plot_bev_lane_det_og_696.png}{-3}{Bev-LaneDet$^{\star}$}
\drawQualitative{overlay_latr_star_696.png}{3d_plot_latr_star_696.png}{xy_plot_latr_star_696.png}{-2}{LATR$^{\star}$}
\drawQualitative{overlay_bev_lane_det_tco_1696.png}{3d_plot_bev_lane_det_tco_1696.png}{xy_plot_bev_lane_det_tco_1696.png}{-1}{Bev-LaneDet~\cite{wang2023bev}}
\drawQualitative{overlay_latr_og_696.png}{3d_plot_latr_og_696.png}{xy_plot_latr_og_696.png}{0}{LATR~\cite{luo2023latr}}
\end{tikzpicture}
\fi 

\includegraphics[width=\columnwidth]{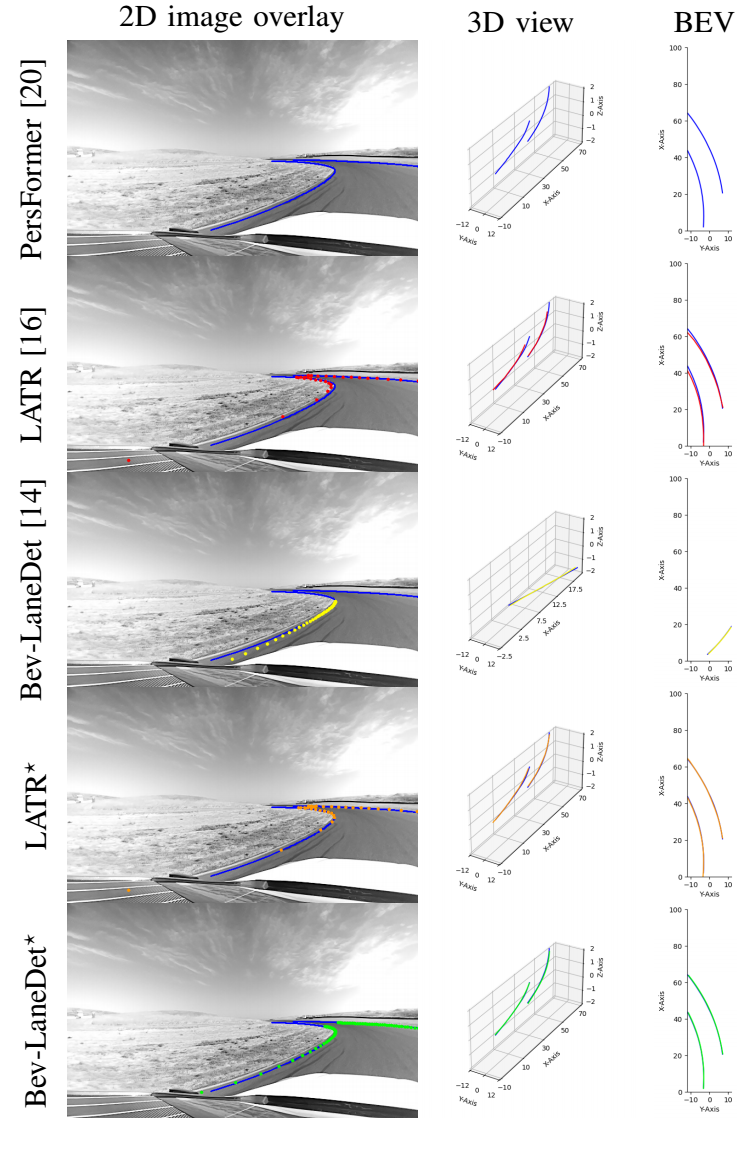}
\vspace*{-20pt}

\caption{Qualitative example of PersFormer, LATR, and BevLaneDet with and without modifications, trained on a subset of \trilanes, only visualizing camera 0. PersFormer fails entirely, while original Bev-LaneDet (yellow) predicts one lane. The unmodified LATR (red) predicts both lanes with slight errors in the $y$-direction, in particular near the end of the lane. The modified versions of LATR$^{\star}$ and Bev-LaneDet$^{\star}$ (green and blue) predict the lanes more accurately in this example (see BEV view).}
\label{fig:qualitative2D3Dsecond}
\end{figure}

\subsection{Qualitative examples}\label{app:qualitative}
To complement the quantitative results in~\cref{sec:results}, we provide qualitative results from the evaluation, representative of the performance in the real-time deployments.

\textbf{Predictor comparison.} To illustrate what the differences in the metrics of~\cref{tab:testTHdata}, we show the predictions of the 3D lane detectors on camera 0 of \trilanes, including 2D, 3D and BEV perspectives with labels and predictions.

%\textbf{Uncertainty estimates.}
%\todo{Add more 2D + 3D views to show what it looks like when we drive around the track.}

In the first example,~\cref{fig:qualitative2D3Dfirst}, PersFormer fails to generate any lane predictions, while LATR, Bev-LaneDet, and LATR$^{\star}$ only predict one of the lanes. The reason that the LATR variants fail to predict the lane in this instance is unclear, but for Bev-LaneDet, the reason is primarily that only one of the lane labels exists in the camera-aligned BEV perspective. In contrast, the modified Bev-LaneDet$^{\star}$ predicts both lanes with high accuracy, and the ``near'' predictions are visibly better than those in the unmodified Bev-LaneDet model.

In the second example, ~\cref{fig:qualitative2D3Dsecond}, PersFormer once again fails to generate predictions, and the unmodified Bev-LaneDet predicts a small portion of one of the lane boundaries, just as in the previous example. LATR, LATR$^{\star}$, and Bev-LaneDet$^{\star}$ all predict two lane boundaries. The prediction is visibly worse with the unmodified LATR when looking at the BEV perspective, and both the modified LATR$^{\star}$ and Bev-LaneDet$^{\star}$ produce good estimates of the lane boundaries. Notably, despite being masked out, the LATR variants tend to extrapolate the lane boundaries in the portion of the image that is occluded by the vehicle. Bev-LaneDet, on the other hand, generally does not predict the invisible portions of the lane boundaries.

\begin{figure}[h!]
    \centering
    \includegraphics[width=\columnwidth]{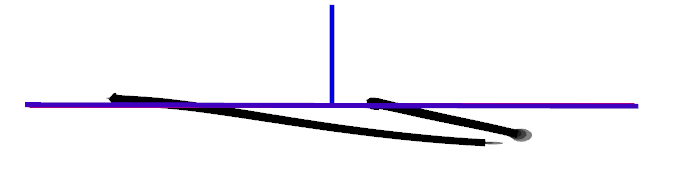}
    \includegraphics[width=\columnwidth]{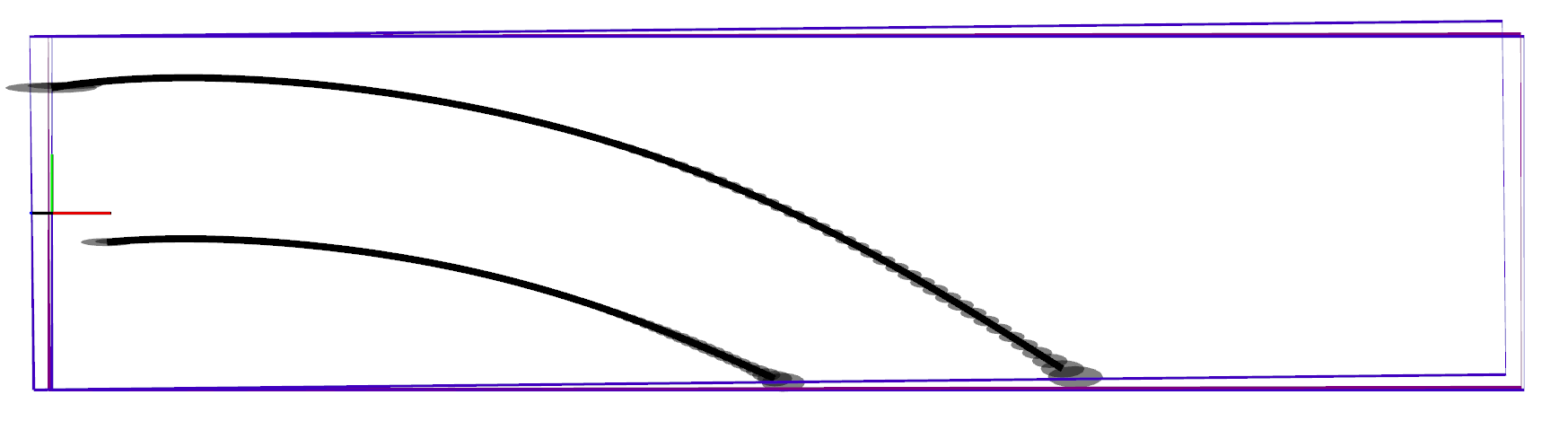}
    \includegraphics[width=\columnwidth]{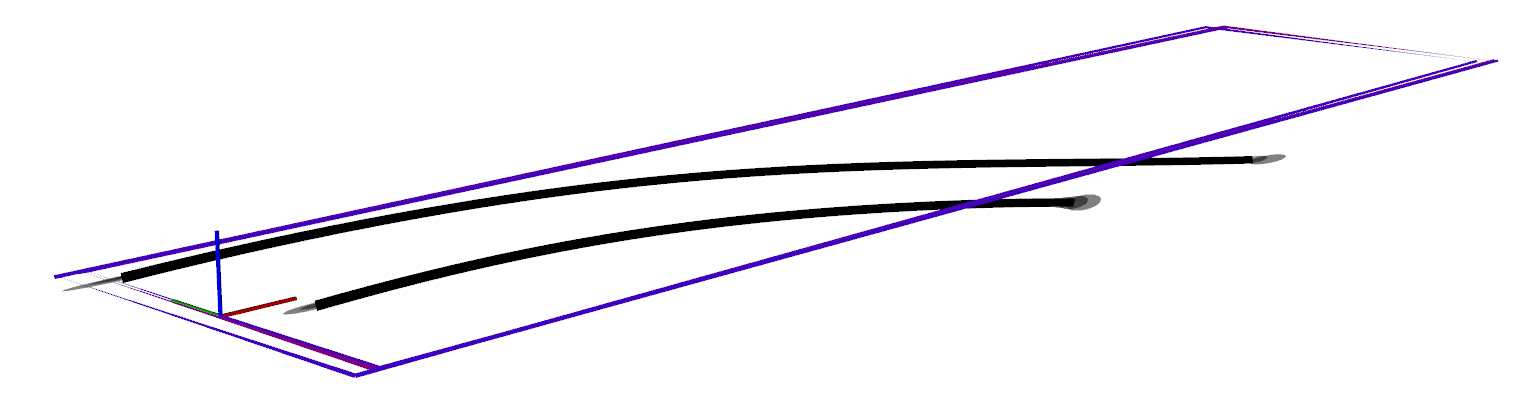}
    \caption{Views of a regressed output with significant elevation changes.}
    \label{fig:qualitativeA}
\end{figure}

\textbf{Elevation variability.} A key consideration for using the 3D detectors to predict lane geometry in the BEV space is the variable elevation along the racetrack. This is captured in~\cref{fig:datadistributions}, and in predictions of the road geometry leading into turn 8, visible in~\cref{fig:racetrack} around (-780,350). Here, the regressed output results in a curve consistent with the ground truth and drops by almost 4m over approximately 50m away from the car (see~\cref{fig:qualitativeA}). A 2D image space predictor leveraging an IPM with a flat ground assumption would not be able to capture such changes in elevation, necessitating an approach that estimates depth~\cite{yan2022once}, or elevation in a BEV perspective~\cite{wang2023bev}, as done in this paper.

%\fi
%\input{7_new_appendix}

\end{document}